\documentclass[10pt]{article}

\usepackage{mathrsfs} 

\usepackage[]{graphicx}    
\usepackage[gen]{eurosym}
\usepackage{verbatim}
\usepackage{amsthm}
\usepackage{amsmath}
\usepackage{dsfont}
\usepackage{amssymb}
\usepackage{latexsym}		
\usepackage{tabularx}
\usepackage{setspace}
\usepackage{listings}
\usepackage{stmaryrd}
\usepackage{hhline}
\usepackage{booktabs}
\usepackage{caption}
\usepackage{bbm}
\usepackage{arxiv}

\usepackage[utf8]{inputenc} 
\usepackage[T1]{fontenc}    
\usepackage{hyperref}       
\usepackage{url}            
\usepackage{booktabs}       
\usepackage{amsfonts}       
\usepackage{nicefrac}       
\usepackage{microtype}      
\usepackage{natbib}

\usepackage[capposition=top]{floatrow}
\usepackage{multirow,lscape,longtable}

\newcommand{\tabitem}{~~\llap{\textbullet}~~}

\title{Applying Machine Learning to Life
Insurance: some knowledge sharing to
master it}

\author{
 Antoine Chancel \\
  Knowledge Team \\                         
  SCOR\\
  Paris, France\\
  \texttt{achancel@scor.com} \\ 
   \And
  Laura Bradier \\
  Knowledge Team \\
  SCOR\\
  Singapore\\
  \texttt{lbradier@scor.com} \\
  \And
 Antoine Ly \\
  Knowledge Team \\                         
  SCOR\\
  Paris, France \\
  \texttt{aly@scor.com} \\
   \And
 Razvan Ionescu \\
  Knowledge Team \\                         
  SCOR\\
  Paris, France\\
  \texttt{rionescu@scor.com} \\ 
   \And
  Laurène Martin \\
  Paris, France  \\
  \texttt{laurene.martin@ensae.fr}  \\
  \And
   Marguerite Saucé \\
  Paris, France\\
  \texttt{marguerite.sauce@gmail.com} \\ 
}

\begin{document}
\maketitle


\keywords{Survival Modelling \and Machine Learning \and Life Insurance \and Python \and Statistical Modeling}

{\Large\textbf{General introduction}}

\section{Motivation and scope of this paper}

Machine Learning permeates many industries, which brings new sources of benefits for companies. However, within the life insurance industry, Machine Learning is not widely used in practice as over the past years statistical models have shown their efficiency for risk assessment. Insurers may thus face difficulties assessing the value of the artificial intelligence.\\

Focusing on the modification of the life insurance industry over time highlights the stake of using Machine Learning for insurers and benefits that it can bring by unleashing data value.\\

This paper reviews traditional actuarial methodologies for survival modeling and extends them with Machine Learning techniques. It points out differences with regular machine learning models and emphasizes the importance of specific implementations to face censored data with the Machine Learning models family. In complement to this article, a Python library has been developed. Different open-source Machine Learning algorithms have been adjusted to adapt the specificities of life insurance data, namely censoring and truncation. Such models can be easily applied from this SCOR library to accurately model life insurance risks. 
This library is briefly presented in section \ref{library}




\subsection{Why consider Machine Learning for risk modeling?}

The life insurance industry became more complex as the modern life insurance industry now offers all kinds of products covering death events or other hazards related to the health condition of the insured. The main products are covering \textit{Mortality}, \textit{Critical Illness}, \textit{Disability}, \textit{Medical Expenses}, and \textit{Longevity}. Besides, thanks to technological advances and data storage capacity improvements, information considered for risk assessment has increased a lot and is still increasing. For instance, nowadays, life insurance applicants are asked to share, besides their age, part of their medical history, financial situation, and their profession. 
In addition to this new information, thanks to the seniority of the industry, insurers may also have access to centuries of data accumulation to deepen their knowledge.\\
As the information available is increasing over time, new challenges for insurance companies and actuaries are arising, namely efficiently extracting and analyzing information from a large amount of data to assess risk better.

\begin{itemize}
    \item Machine Learning as a solution to deal with huge databases
\end{itemize}
For many years, only a limited amount of information about the applicants was collected. For instance, at a time, only age, gender, and smoking status of an insured were used to assess some biometric risks. Therefore, simple models such as linear regressions or classifications were considered sufficient to grasp the risk.\\
However, by using such simple models the potential of the current increasing amount of data in the insurance industry may not be optimally tapped into.\\Indeed, insurers are handling larger and larger databases, both \textit{vertically} (very large amount of policies sold) and \textit{horizontally} (numerous  features collected for an insured). This may lead to difficulties in the calibration of simple models such as linear regressions.
\begin{itemize}
    \item And to tackle new types of data
\end{itemize}
In addition to collecting more data, insurers are also collecting new types of data that cannot be handled by traditional statistical models. This is the case for textual data, or badly structured data, such as large databases that integrate correlated and non-linear relationships between variables. \\ 
One can also mention real-time data collected on distributed storage systems in the cloud via connected objects. The number of steps in a day is an example of such real-time data. 
\\ \\
Thanks to the increased computing power, the insurance industry can use Machine Learning models to capture the increasingly complex information contained in these new and larger datasets.

\subsection{Which insurance departments can benefit from Machine Learning risk modeling?}
By leveraging on the increasing amount of data collected, Machine Learning allows insurers to refine their risk modeling and understanding.\\
This benefits all insurance departments such as - to name a few -
\begin{itemize}
    \item \textbf{Underwriting}:\\
    A better understanding of the main risk drivers allows a thinner assessment of the applicants' risk profiles and therefore a better accuracy in the acceptance, rejection or loading granted.\\ In addition, the newly collected data can have a really good predictive power while being less intrusive for the applicant who provide it or less expensive/easier to access to (for instance, asking for the age of an applicant is easier than asking for blood measurements). This leads to an enhancement of the customer journey.
    \item \textbf{Pricing}:\\
    Machine Learning modeling can be directly used to predict incidence/morbidity rates. Some insights from the modeling such as variable importance can help decide on the pricing risk factors to use (provided that they are in line with local regulations and ethics).
    \item \textbf{Experience Analysis}:\\
    By being a tool to study complex data (correlations, ...), Machine Learning techniques allow to highlight new insights from the experience data and thus to better understand risk drivers. In particular, partial dependence, variable importance or SHAP graphs (that are underlying components of Machine Learning) are powerful tools to precisely understand risks.\\
    Such insights can then be considered to better monitor the in-force portfolios.
    \item \textbf{Product development}:\\
    Machine Learning allows the use of new types of data for risk modeling and therefore the consideration of new insurance products. For instance, this is the case for the SCOR BAM products using continuous physical activity data to revamp mortality and critical illness products, making them more inclusive and simplifying the customer journey.
\end{itemize}

\newpage 

\section{Modeling risk in life insurance}

Biometric risks require to model the duration until the occurrence of an event.\\
Such an event depends on the insurance product and what it covers.\\
For instance:
\begin{itemize}
    \item \textbf{Mortality / Longevity products} require to model the lifespan which is the duration before the death of the insured,
    \item \textbf{Disability products} require to model:
        \begin{itemize}
            \item For the incidence rates assessment: the duration until the occurrence of disability
            \item The termination rates assessment: the disability duration which is the period an insured will remain disabled i.e. duration before death or recovery 
        \end{itemize}
    \item \textbf{Critical Illness products} require to model the duration before the occurrence of a covered condition
    \item \textbf{Long Term Care products} require to model: 
        \begin{itemize}
            \item The autonomy duration which is the duration for non-disabled individuals before death or the loss of autonomy
            \item The non-autonomy duration which is the duration for a disabled (non-autonomous) individual before death (we do not consider recovery)
        \end{itemize}
\end{itemize}
The underwriting field of a life insurance company aims at classifying insurance applicants from \textit{bad} to \textit{good} risk based on their risk profile. That means being able to rank applicants based on their probability of claim occurrence (and claim duration for risks with a termination component such as disability or long term care). Besides, the development of more precise models contributes to being more inclusive, as we may derive a price even for a very risky individual. Thus it enables to sell products to clients who would have been excluded from the portfolio in the past.\\
For the computation of premiums and reserves, one may not limit the modeling to a binary classification, indicating whether the event has occurred or not. Indeed, it is important to be able to compute the event occurrence probability at any time for the whole duration of the contract. In other words, we seek to predict the event probability for a given period of time. \\ \\
Predicting time to event requires a specific modeling approach called \textit{Survival Analysis} mainly due to the underlying data structure. Survival Analysis is a branch of statistics for analyzing the duration until one or more events happen. Formally, it is a collection of statistical techniques used to describe and quantify the time to a specific event such as death, disease incidence, termination of an insurance contract, recovery, or any designated event of interest that may happen to an individual. The time may be measured on different scales such as years, months, or days from the beginning of the follow-up of an individual until the occurrence of the event. The time variable is usually referred to as survival time because it gives the time that an individual has “survived” in his/her initial state over some follow-up period.

\newpage

\section{Where can I find data to model duration?}

Survival datasets are precious for the life insurance industry and have many purposes, such as refining risk knowledge, making life insurance more inclusive, supporting the development of new products, innovating, ...\\
Such data can be even more important than the model itself when dealing with Machine Learning modeling. In addition, in insurance, data science will value such data even more if it is already correctly integrated in a process.\\
Whether the data used be internal or external, the \textbf{quality of the data} should be correctly assessed before starting a project and collecting business requirements. Indeed, the quality of the solution that will be created is directly driven by the quality of the data. A good understanding of the business and the data is therefore crucial and should be reviewed by a qualified actuary.

\subsection{Internal data}

Insurance data is the aggregation of all the data that the company possesses related to the business: claims, underwriting questionnaires, portfolio follow-ups, ... Those data sources are frequently separated in different systems, locations and environments. If necessary, the \textbf{consolidation} of those systems is the very first step of an internal project. This step can be time consuming but the performance of the models will heavily depend on the quality of the data preparation .\\
Note that this stage is sometimes not even possible. This is the case, for instance, if a primary key is missing or not recorded to link the information between different databases or tables. Proxies between common variables can be proposed but the success of the project can be compromised.

\subsection{External data}

Public institutions such as hospitals, universities, governments, ... may record relevant medical data about the population as well as its health status. More recently, innovative technological companies have started collecting new features such as physical activity or genomics to predict the life expectancy of a person or the risk of certain illnesses (cancer, heart attack, stroke, ...).\\
Insurers and reinsurers are more than ever willing to build partnerships with those companies to collaborate on the development of new innovative models.

\section{What is the specificity of survival data?}

Most of the time, survival duration is only partially observed. Because of it, standard Machine Learning approaches can not be directly applied. \\
In general, survival data contain a distinctive characteristic making it impossible to directly measure survival. Indeed, while gathering the information, undesired events can occur during the observation period that pollute the records and prevent the observation of the full survival duration for all individuals. One may think of events such as Lapses, Hospital transfers, IT system failures during recordings, etc. Besides, the observation time is limited, so the event of interest, such as death, may occur outside of the time window. These characteristics are called \textbf{Censoring} and \textbf{Truncation}. 

\subsection{Censoring and Truncation}

Censoring defines a situation in which the information is only partially known or observed, while Truncation corresponds to a situation in which the information is totally unknown. There are two types for each: Right and Left.

\begin{figure}[H]
    \centering
    \includegraphics{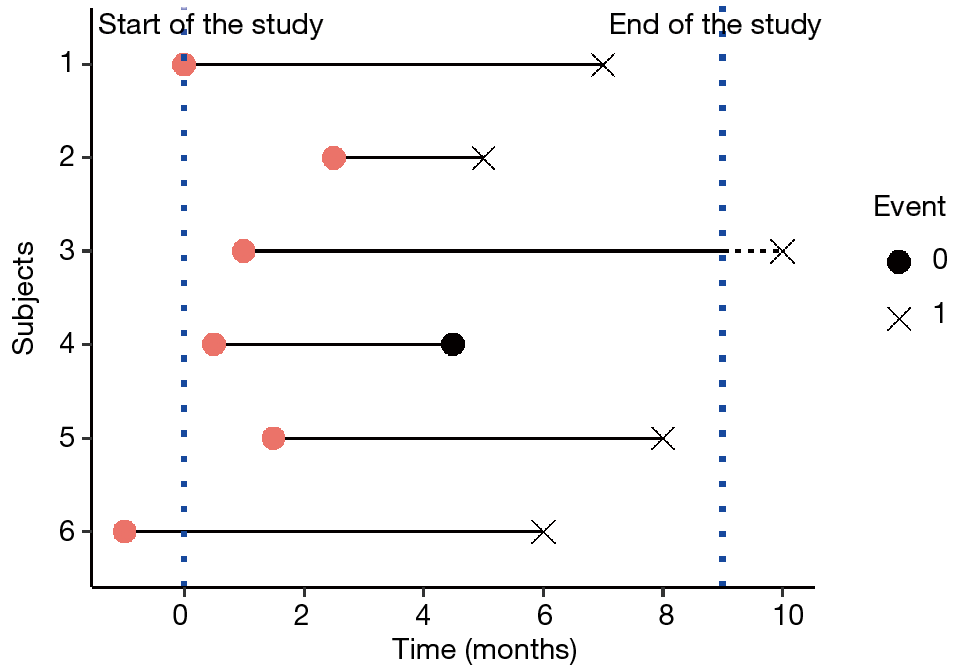}
    \caption{Illustration of censoring.}
    \label{cen}
\end{figure}

\begin{itemize}
    \item[$\bullet$] \textbf{Right Censoring}: Right censoring refers to an event that occurs after the end of the observation period or to the loss of the trace of a subject due to other independent reasons. In Figure~\ref{cen}, subjects 3 and 4 are subject to right censoring. Even if the exact duration is not observed, right censoring still reveals partial information: the event of interest occurs after the observed time and thus the life duration is at least as long as the censoring duration.\\
    Let's consider the study of a life duration after the purchase of a life insurance contract. If the insured ends her/his contract $t$ years after the subscription, then for this observation, we can only say that the insured survived at least $t$ years. But her/his survival time, i.e. period until death occurs, is unknown.
    \item[$\bullet$] \textbf{Left Censoring}: Left censoring is quite the opposite of right censoring. It occurs when the trigger point of the duration measure is before the observation period as it is the case for the subject 6 in Figure~\ref{cen}. Once again, we only know that the real duration is longer than the one observed.\\
    One may face left censoring when a study includes individuals who have already a contract at the beginning of the observation time. The purchasing date is thus unknown. When studying life span, left censoring isn't an impediment provided that the date of birth is known. However, if the study is on the duration before lapse for instance, one can see that the left censoring is indeed making the observed information only partial.
    \item[$\bullet$] \textbf{Right Truncation}: Right truncation corresponds to individuals who are completely excluded from a study because the starting event that includes them in the study happens after the end of the observation period.\\
    Considering the same example as before, right truncation is observed when an individual buys insurance after the observation period. He/she is thus totally excluded from the study.
    \item[$\bullet$] \textbf{Left Truncation}: Left truncation is the opposite of Right truncation. An individual is excluded because the event of interest occurs before the beginning of the observation period.\\
    For mortality risk modeling, all individuals who are insured and whose death occurs before the observation period are left truncated and thus excluded of the study. \\
\end{itemize}

Most of the time, when studying biometric risks in life insurance, left censoring and right truncation don't occur.\\
Left truncation is more likely to happen when modeling life risks, but \textbf{hereinafter, we will only deal with right censoring, which is the most common scenario.}\\ It is worth noting that it is possible and quite easy to consider left truncation by enhancing the modeling a bit.

\subsection{Why is considering censoring and truncation important?}

When dealing with survival data, a common mistake could be to simply ignore any censoring or truncation effects (referred to as \textbf{Mistake1} below). This approach would lead to an underestimation of the probability of the event of interest. Another common mistake would be to restrict the study to observations that are complete by removing any censored or truncated records (referred to as \textbf{Mistake2} below). Here as well, the estimation would be extremely biased.\\
Let's consider the four following individuals to understand the intuition behind the importance of taking censoring into account:

\begin{figure}[H]
    \centering
    \includegraphics[scale=0.7]{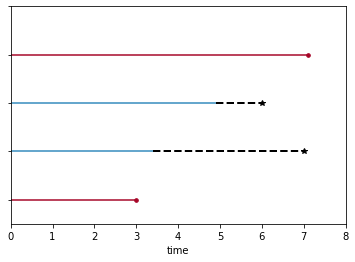}
    \scalebox{0.5}{
    \begin{tabular}{|c|cccc|}
        \hline
        & \textit{\textbf{1}} & \textit{\textbf{2}} & \textit{\textbf{3}} & \textit{\textbf{4}} \\ \hline
        \textit{\textbf{Duration observed}} & 7.1 &  4.9 & 3.4 &  3 \\
        \textit{\textbf{Real duration}} & 7.1 &  6 &  7 &  3 \\
        \textit{\textbf{Status}} & D & C & C & D  \\ \hline
    \end{tabular}}
\caption{Illustration to highlight the impact of censor}
\label{ignor}
\end{figure}

Based on the data illustrated above, not considering censoring appropriately would lead to:

\begin{itemize}
    \item Mistake1: the average survival duration is 4.6 when considering censored time (i.e Duration observed),
    \item Mistake2: the average survival duration is 5.05 when removing the censored observations (i.e removing observations 2 and 3),
    \item While the real average survival duration is 5.775.
\end{itemize}
In both cases of this example, life expectancy is underestimated when the partial information coming from censored individuals is ignored.

\section{SCOR Python survival models library} \label{library}

A library in Python was developed to apply all the presented methods. This library adapts the existing open-source library of Machine Learning and deep learning to our needs in actuarial science. This package was developed  with the Data Analytics team at SCOR and is distributed for all SCOR collaborators. The modules customize some existing statistical and Machine Learning models: sklearn, pygam, statsmodels, lifelines, scikit-survival, lightgbm and catboost.\\
In addition, elements helping the interpretation of models, such as Shapley values, partial dependences, ..., have been adapted to our survival models.

\begin{figure}[H]
    \hspace*{-1cm}
    \includegraphics[scale=0.50,angle=90]{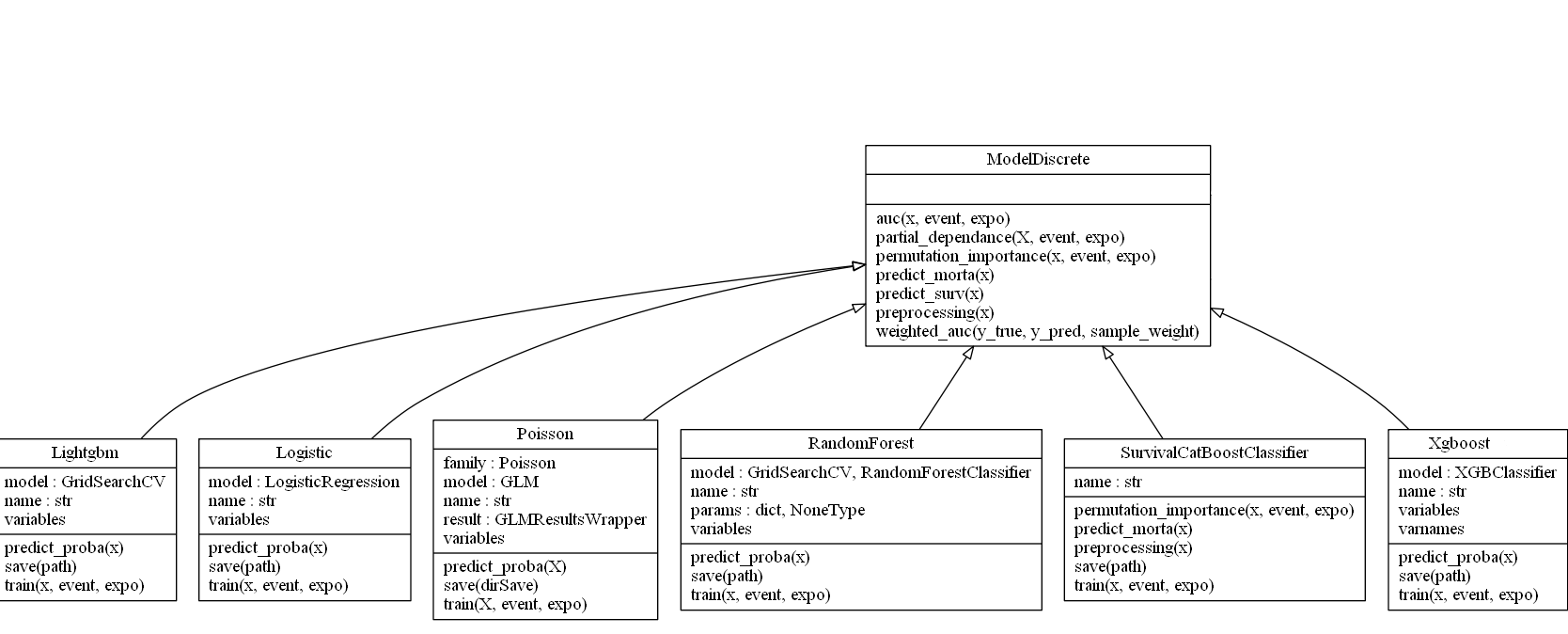}
    \caption{Discrete model classes}
    \label{uml}
\end{figure}

\begin{figure}[H]
    \includegraphics[scale=0.58,angle=90]{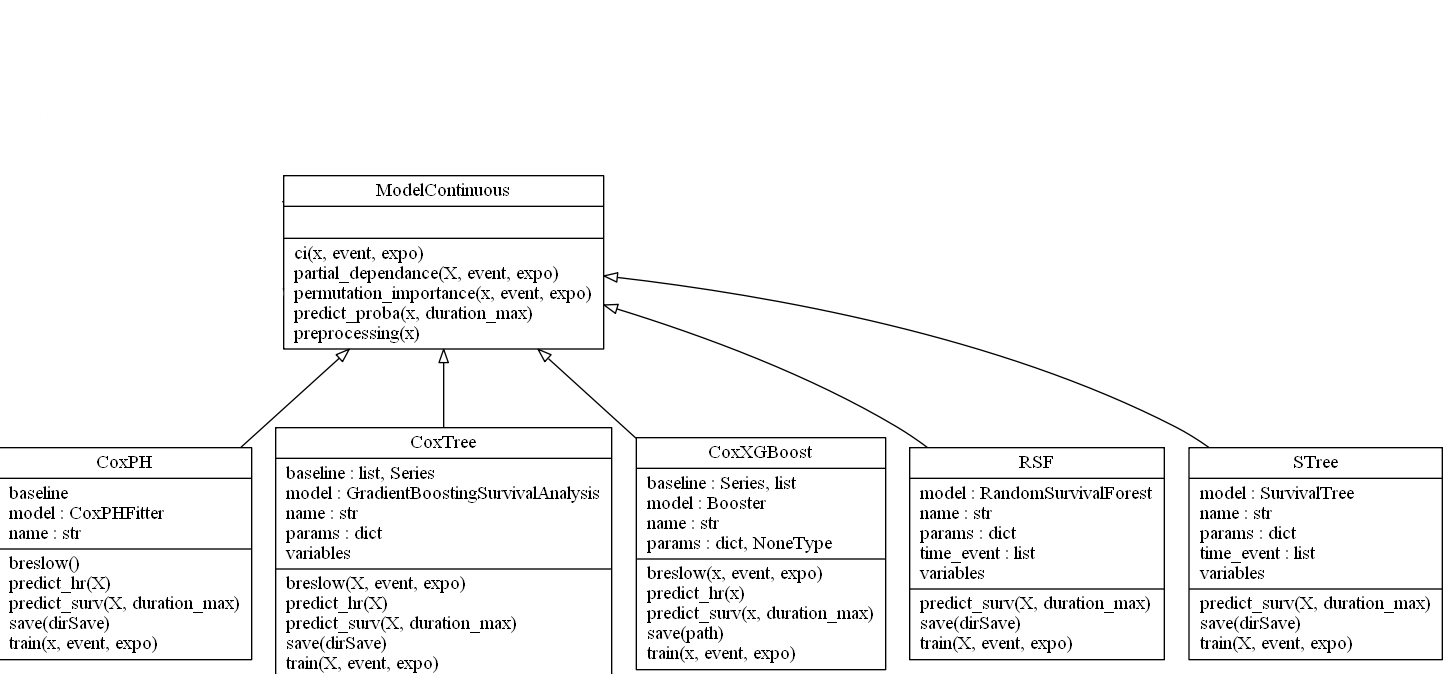}
    \caption{Time-to-event model classes}
    \label{uml2}
\end{figure}
{\Large\textbf{Predicting life duration in practice}}

This section provides a high level overview of techniques to model duration and is addressed to non-expert readers. Deeper explanations for each step of the process are available in the following sections.\\
We will focus on the use of the methods implemented within the Python library and their interest on a business matter to automate and facilitate the mortality modeling of an insurance portfolio. The calculations are performed on the NHANES database as it is an open-source mortality experience database that contains a significant amount of information.

\section{Dataset presentation}
\subsection{NHANES overview}
NHANES is the acronym for National Health and Nutrition Examination Survey, which is a study program designed to assess the health and nutritional status of adults and children in the United States. In this program, information is collected through interviews and examinations.\\
Interviews gather declarative information about a participant's lifestyle (e.g. drinking and smoking habits) and self-assessment of health status. The information from such interviews is rather qualitative.\\
For their part, physical examinations allow to collect objective quantitative information about a participant's characteristics and health status - such as height, weight, blood pressure or cholesterol, to name a few.\\
Ultimately, more than 5,000 variables are recorded and they can be categorized into 5 classes: demography, dietary, laboratory, examination and questionnaire.\\
Besides these different risk factors, the observed duration in months is indicated for each individual. An indicator variable specifies whether the end of the observation period is due to the death or to another independent event. \\
The following figure illustrates the structure of the NHANES dataset for six individuals with the extraction of eight risk factors.

\begin{figure}[H]
    \centering
    \includegraphics[scale=0.5]{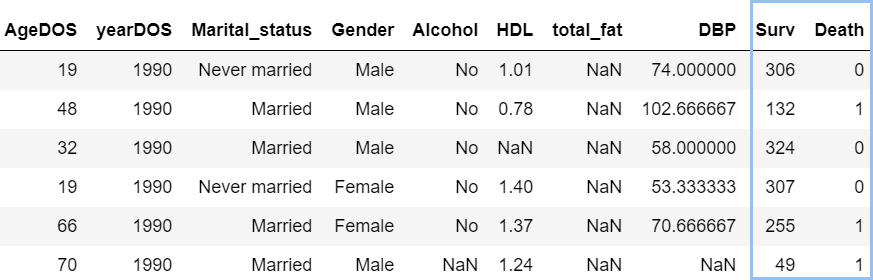}
    \caption{NHANES information extraction}
    \label{data1}
\end{figure}

\subsection{NHANES subset used for our study}
The aim of our study is to predict the duration before death from multiple risk factors (using observation duration and death indicator variables available for each individual).\\
To that extent, we initially used a subset of the entire NHANES dataset which included more than 65,000 observed individuals and 106 explanatory variables. 
Then, following our analysis, we only consider the most meaningful twenty-nine variables described in Appendix~\ref{ch:NHANES} (in addition to the duration and death indicator). Some of the variables indeed contain too many missing values or redundant information. Therefore we excluded them. \\

The NHANES program aims at being representative of the whole non-institutionalized US population. To that extent, a weight is associated to each person studied in order to readjust the sample to the non-institutionalized national population. \\
\textbf{In the following, we will ignore the weights and consider each line as an individual.}\\
The purpose of our study is to highlight how some factors may impact mortality. We do not have an interest in the representativeness in terms of the American population, therefore it is acceptable to ignore the weights.\\
However, we want the insights of this study to be valid in a life insurance context. Consequently, and based on the available variables, we try to replicate a simplified life insurance underwriting process to create a fictive life insurance portfolio: only participants who would have been accepted upon life insurance application have been kept. For instance individuals with too high BMIs or some pre-existing conditions were excluded. \\

Finally $29,870$ individuals are available to derive mortality patterns based on twenty-nine risk factors.

\section{Data pre-processing}

The quality of the data is essential to any project. As such, we begin by pre-processing the data.

\subsection*{How should I tackle missing information?}

The proportion of missing values is quite significant for some of our variables, which means that we have to apply missing value imputation to properly train the models. Two approaches have been considered, depending on the variable type.\\
For numerical variables, the median of the series was assigned to all unspecified values.\\
For categorical variables, a new category was created for each variable as \textit{'Missing value'}. This might enable us to highlight potential patterns specific to individuals with a given missing variable.\\
From raw data~\ref{data1}, we get the following complete table:

\begin{figure}[H]
    \hspace*{0.9cm}
    \includegraphics[scale=0.8]{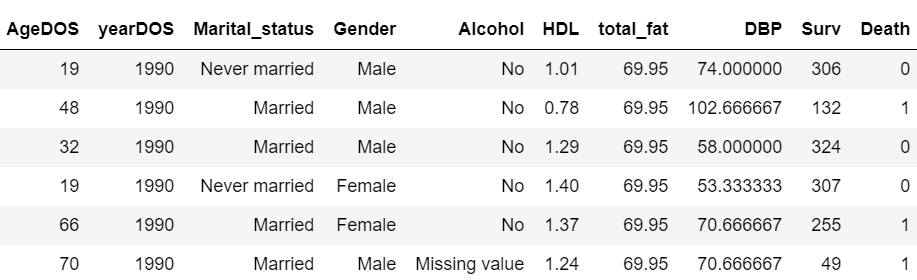}
    \caption{Missing value imputation illustration}
    \label{data2}
\end{figure}

\subsection*{How can I include categorical variable into models?}

As some models cannot deal with categorical features, we apply one-hot encoding to all factor variables. One-hot encoding is a representation of categorical variables as binary vectors. This method produces a vector with length equal to the number of categories in the data set. If a data point belongs to the $i^{th}$ category, then components of this vector are assigned the value 0 except for the $i^{th}$ component, which is assigned a value of 1. In this way, one can keep track of the categories in a numerically meaningful way.\\
Figure~\ref{data3} illustrates the process for the marital status variable of the raw table~\ref{data1}:

\begin{figure}[H]
    \hspace*{-1.1cm}
    \includegraphics[scale=0.7]{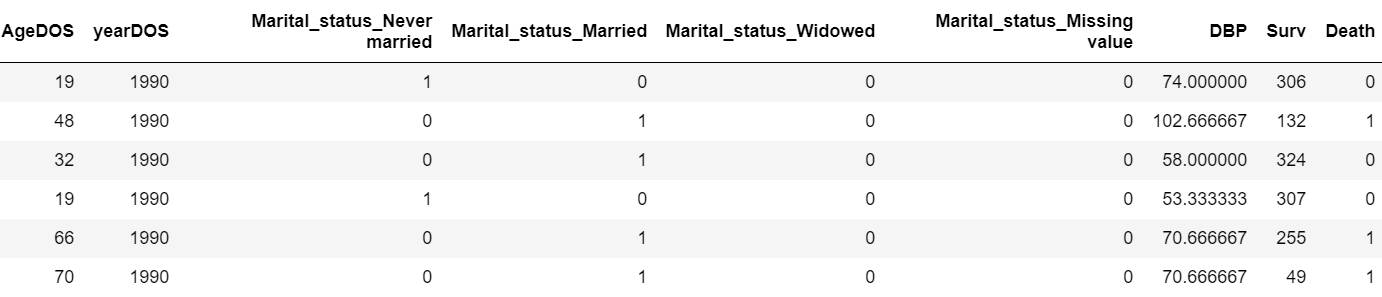}
    \caption{One-hot encoding illustration}
    \label{data3}
\end{figure}

\subsection*{Creation of the "Pseudo data tables"}

Pseudo data tables are detailed in \textbf{section \ref{sec:pseudo}}.
In the original dataset, the follow-up time in months, as well as a death indicator, are available, which corresponds to a traditional survival data table.\\ However depending on the modeling strategy (\textbf{cf Section~\ref{sc:discrete}}), using pseudo data tables may be necessary. It consists in discretizing the data and computing exposures to risk. Such a transformation of the data into pseudo survival data tables is an important step in the pre-processing of survival data.  \\
In our study, we consider the survival time in years rather than in months. A yearly basis seems granular enough for life insurance mortality modeling purposes. Besides, this choice enables us to considerably reduce the number of rows in the discetization and thus the computation time.\\
Considering the approach described in \textbf{section~\ref{sec:pseudo}}, for each observation (i.e. individual) we create as many rows as the number of years the individual was observed for. For each one of the created rows, we add an explanatory variable: \textbf{duration}.\\
"Duration" represents the time interval for which exposures (\textit{the initial} and \textit{the central}) are computed. \\
The final step is to modify the time-varying variables. In this case, age and year are the only time-varying variables. Thus, we only have to increment them by one in each interval. This means computing the \textbf{Attained Age} (respectively {\textbf{Attained Year}}) by summing up the age (respectively year) at the start of observation and the yearly duration.\\
Note that in order to ease calculations, we assume the same day and month for the date of birth and the date of observation.\\

Within our SCOR survival library, a specific class has been created to transform of a survival data table into a pseudo data table: 

\begin{lstlisting}[language=Python]
from scor_survival.discretizer import exposure_expansion
pseudo_data = exposure_expansion(survival_data,
                                 time, event, individual_key,
                                 entry_age, entry_year)
\end{lstlisting}

The application of this function to the last record of the data in Figure~\ref{data1} (seventy-year-old married man) produces the following output:

\begin{figure}[H]
    \hspace*{0.2cm}
    \includegraphics[scale=0.6]{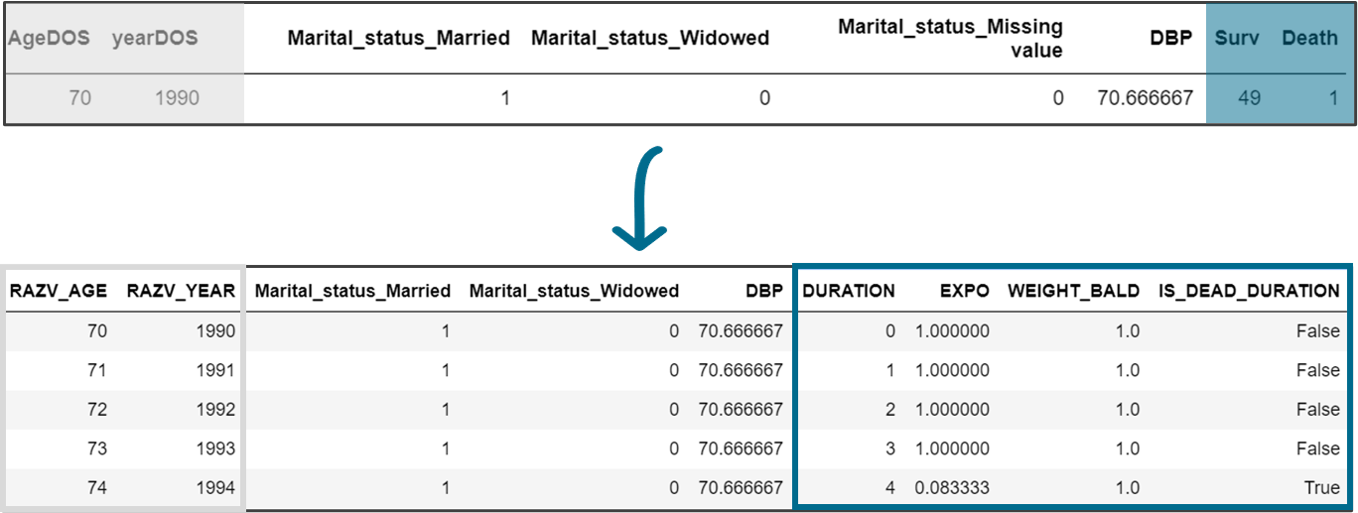}
    \caption{Transformation to pseudo data table : illustration}
    \label{data4}
\end{figure}

\subsection*{Splitting the data into train and test sets}
Before calibrating different models, we split our data into a train and a test sample by making sure the split is stratified and the mortality contained within the two subsets is equivalent to the global mortality of 9\%.

\section{Model calibration}

In data science, the calibration of model hyper-parameters is essential as it impacts the model performance. It is advised to use a separate hold-out to perform the research of hyper-parameters with k-fold cross-validation. \\
The calibration may be challenging as it often implies long and tedious calculations. This is especially the case for tree-based models (hyper-parameters are \textit{depth, leaf sizes, estimator numbers, learning rate, etc.})\\
The lack of computation resources to train a model may lead to a too tiny grid of parameters to test (for instance, when an XGBoost is trained with an early stop on the number of estimators). This could lead to over-fitting the model, which users should be aware of, with the risk of reaching local optimums (while testing larger ranges of parameters minimizes this risk).

\begin{figure}[H]
    \hspace*{-2.1cm}
    \includegraphics[scale=0.95]{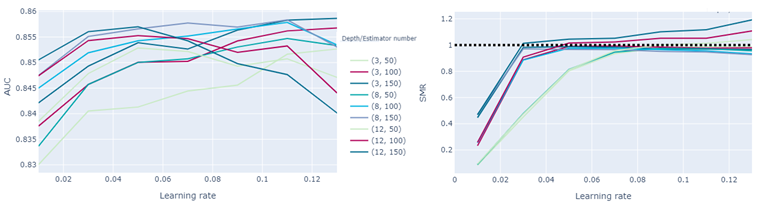}
    \caption{Catboost hyper-parameters selection}
    \label{gs}
\end{figure}

Based on this thought, within the library, every model is extended to implement a \textit{scikit-learn} interface. This allows parametrization before training a model, and to keep the standard syntax commonly used by Python users. 
A default grid of parameters is also suggested as a first very parsimonious model, based on SCOR experience. Thus, it eases the implementation of survival models by using default settings that minimize risk of mis-parametrisation. 
As illustrated in Figure~\ref{gs} for the CatBoost model, the parameter choices imply very different performance results. We applied a grid search method to calibrate them based on two performance metrics.\\
Based on the AUC (cf Section~\ref{sec:AUC}), the best model is obtained for a fit with 150 estimators of depth 8.\\
To keep the SMR close to 100\% (cf Section~\ref{sec:SMR}), we retained a learning rate of 0.07. This calibration appears to us as a good balance, as it provides good results on both metrics considered, AUC and SMR.\\
\begin{lstlisting}[language=Python]
cb = scor_survival.models.discrete.SurvivalCatBoostClassifier()
g = {'depth':[3,8,12], 'learning_rate':np.arange(0.01, 0.15, 0.01), 'iterations':[50,100,150]}
gs = cb.gridsearch(g, X, y, w)
\end{lstlisting}

To avoid overfitting, it is important to tune a model with a similar AUC performance on the train and test datasets. The presence of over-fitting is highlighted in the figure, as for many deep trees and high learning rates, the AUC is decreasing on the test set.\\
All the models presented in the following have been calibrated through a k-fold cross-validation grid search based on the weighted AUC (respectively the C-Index (cf Section~\ref{sc:CI})) for discrete models (respectively continuous models). 

\section{How can I ensure that a model is well fitted?}

A survival model is an estimator that should respect certain mathematical properties. The complexity of the models can lead to biased or inadequate models.\\
Using inaccurate models in insurance is a threat to the business. Constant monitoring of the development process and the usage of the model is mandatory to reduce the risk.\\
Our main focus is to obtain a non-biased model. Mathematically, the non-bias property for parametric models is:
\begin{equation}
     E_X[E_Y[Y|X]] = E_Y[Y]
\end{equation}

This equation states that the weighted average of the predictions is equal to the actual risk measured on the data. The model is expected to replicate the average mortality of the dataset. This equation consideration should always hold on the training set. This is the very first test to conduct after the calibration of a model.\\
Within the library developed, a function implements this test for all continuous (cf Section~\ref{sc:continuous}) and discrete models (cf Section~\ref{sc:discrete}).
\subsection{Continuous models}
\begin{figure}[H]
    \hspace*{1.7cm}
    \includegraphics[scale=0.72]{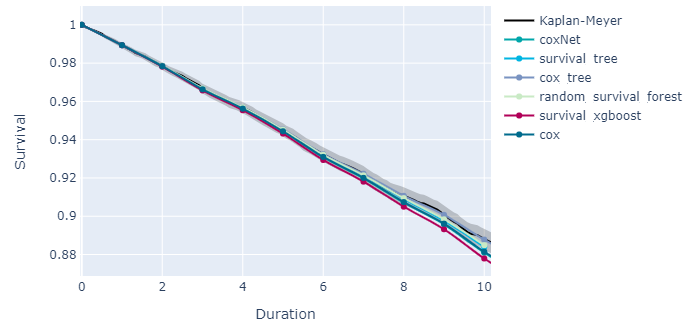}
    \caption{Validation for continuous model}
    \label{sc}
\end{figure}

For continuous models, this test is the comparison of the average of the survival curves predicted by the model with the Kaplan-Meier (cf Section~\ref{sc:KM}) estimation (cf Figure~\ref{sc}).\\
Even if it seems that on average the mortality is a little overestimated by our models for the high duration, we can still consider that all our models are acceptable. Only the Cox-XGBoost model is not contained within the Kaplan-Meier confidence interval for the last duration.\\
The increase of the gap (predicted survival VS Kaplan-Meier survival) over duration is expected. Indeed, the survival curve being cumulative, small errors from previous durations are added successively. \\
To go beyond graphical validation, one can opt for statistical tests to determine whether two survival curves (here the average of the survival curves predicted VS the Kaplan-Meier survival curves) can be considered identical or not. Such tests include but are not limited to the Log-Rank test, the Wilcoxon test, or the Kolmogorov-Smirnov test.\\
Note that specific care is needed when comparing crossing survival curves. Indeed, in this context, traditional tests may lead to misleading results and thus alternative statistical tests should be considered (eg. weighted Log-Rank test or modified Kolmogorov-Smirnov test to name a few). 

\begin{lstlisting}[language=Python]
from scor_survival import analysis
models = [coxnet, stree, coxtree, rsf, cox_xgboost, cox]
analysis.plot_average_survival(models, X, expo, event, max_duration)
\end{lstlisting}

\subsection{Discrete models}
For discrete models, the validation test consists in computing the ratio of the number of observed deaths to the number of deaths predicted by the model: this is the SMR.

\begin{table}[H]
\centering
\scalebox{0.8}{
\begin{tabular}{|r|cc|cc|}
\hline
\textbf{Models} & \textit{SMR Train} & \textit{SMR Test} \\ \hline
\textit{\textbf{Binomial Regression}} & 0.99 & 0.99  \\
\textit{\textbf{Poisson Regression}}  & 1.00 & 0.98  \\
\textit{\textbf{logistic GAM}}        & 1.00 & 0.96  \\
\textit{\textbf{Random Forest}}       & 0.99 & 0.99  \\
\textit{\textbf{LightGBM}}            & 0.99 & 0.98 \\
\textit{\textbf{XGBoost}}             & 0.99 & 0.94  \\
\textit{\textbf{Catboost}}            & 1.01 & 0.99  \\ \hline
\end{tabular}}
\caption{Validation metrics for discrete models}
\label{ev}
\end{table}

The SMR must be close to 1 on the train set if the model is well trained.\\ Based on the table above, all our models seem acceptable as the SMR on the train and test sets are close enough to 1. A deeper comparison will be conducted to define which model has the best predictive performance.\\ 
However, another criteria to consider is the difference in SMR between the train and the test sets. If such a difference is significant, this highlights that the model is over-fitting. \\
The $XGBoost$ model appears to be less effective than the other ones as it highly relies on the hyper-parameter choice. As it requires more computation capacity than the others, it is quite difficult to test a large grid of hyper-parameters. \\
The logistic GAM model seems to be over-fitting as well. Indeed, the model fits perfectly the train set but is biased on the test set. As the standard error on the test set is equal to 5.3\%, this bias on the test set can be acceptable (as a perfect accuracy of 100\% still belongs to the confidence interval of the SMR on test set)

\section{How can I understand a model?}

\subsection{Partial Dependence}

Visualizing the Partial Dependence is interesting to capture the causal effect of a variable on mortality and neutralize the hidden effects of other variables (cf Section~\ref{sc:pd}).

\begin{lstlisting}[language=Python]
from scor_survival import analysis
model = [logistic, poisson, rf, lightgbm, xgboost, gam, catboost]
analysis.plot_partial_dependence(model, var, X, exposition, event)
\end{lstlisting}
\begin{figure}[H]
    \hspace*{1.9cm}
    \includegraphics[scale=1.1]{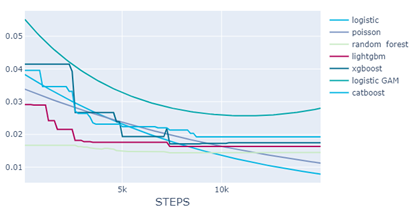}
    \caption{Partial Dependence for discrete models}
    \label{pd}
\end{figure}

The Partial Dependence above showcases a decreasing marginal effect of the number of steps on the predicted mortality (with potentially a flattening of the effect after a steps threshold). It seems that the random forest model is less effective in capturing this trend as its Partial Dependence curve is almost flat. It highlights that the effect seems to be non-linear, which lets us conclude that the mortality will be better-modeled with Machine Learning or non-linear models.

\subsection{Variable importance}


For all our models, the variable \textit{Age} is the one with the highest impact. As the impact of age is way higher than the others, we remove it from the charts for the sake of visibility.\\
In the figures below, the Variable Importance (cf Section~\ref{sc:pi}) is plotted for all our implemented models. We can observe that even if the values are little different from a model to another, the ranking of the variables is quite similar. \\ 
Since the models are fitted on different data, (\textit{pseudo data tables} or \textit{survival data tables} depending if the model is continuous or discrete) two specific figures are provided by type of model.\\ 
Regarding the variable \textit{YEAR}, we can note that it is considered only in the discrete models as duration is embedded in the continuous model.\\

\begin{lstlisting}[language=Python]
from scor_survival import analysis
models = [coxnet, stree, coxtree, rsf, cox_xgboost, cox]
analysis.plot_var_importance(models, X, exposition, event)
\end{lstlisting}

\begin{figure}[H]
    \hspace*{0.3cm}
    \includegraphics[scale=0.9]{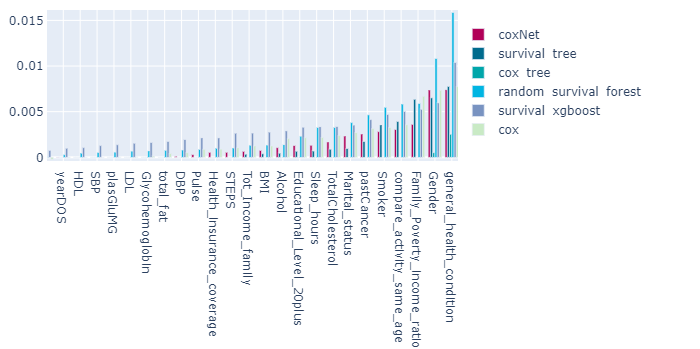}
    \caption{Variable Importance for continuous models}
    \label{varimpC}
\end{figure}

\begin{lstlisting}[language=Python]
from scor_survival import analysis
models = [logistic, poisson, rf, lightgbm, xgboost, gam, catboost]
analysis.plot_var_importance(models, X, exposition, event)
\end{lstlisting}

\begin{figure}[H]
    \hspace*{-0.8cm}
    \includegraphics[scale=0.9]{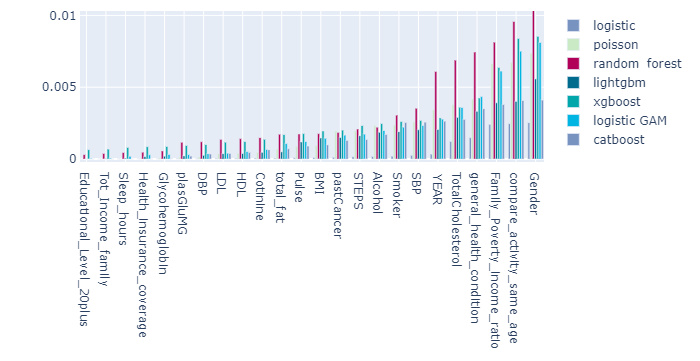}
    \caption{Variable Importance for discrete models}
    \label{varimp}
\end{figure}

For both continuous and discrete models, methods based on the bagging of trees seem to be accentuating the importance of some risk factors compared to other models.\\
Based on the figures, it seems that the highest impact of mortality comes from the socio-economics and general health features, which may be seen through the importance of the family income and the general health conditions and habits, such as being active and consuming alcohol or cigarettes. This phenomenon has been highlighted in some social science studies, such as the one published by~\citet{DeathSocial}, which concludes that the number of deaths attributable to social factors in the United States is comparable to the number attributed to pathophysiological causes.

\section{How can I choose the most suitable model?}

\subsection{Performance metrics}

\subsubsection{ROC Curve}

\textbf{Refer to section \ref{sec:AUC} for details and illustrations on ROC curves and AUC}\\ \\
A comparison of multiple classifiers is usually straight-forward thanks to ROC curves - Receiver Operating Characteristic curve -, especially when such curves are not crossing. Curves close to the perfect ROC curve have a better performance level than the ones close to the baseline. The baseline is achieved by a classifier with a random performance level and corresponds to the straight line from the origin to the top right corner.\\
The area under a ROC curve, which is named AUC, represents an estimation of the performance level. A perfectly predicting model (perfect ROC curve) would have an AUC of 1 whereas a randomly predicting model (ROC curve is the diagonal from origin to top right corner) would have an AUC of 0.5.\\

\begin{figure}[H]
    \hspace*{-2.4cm}
    \includegraphics[scale=0.6]{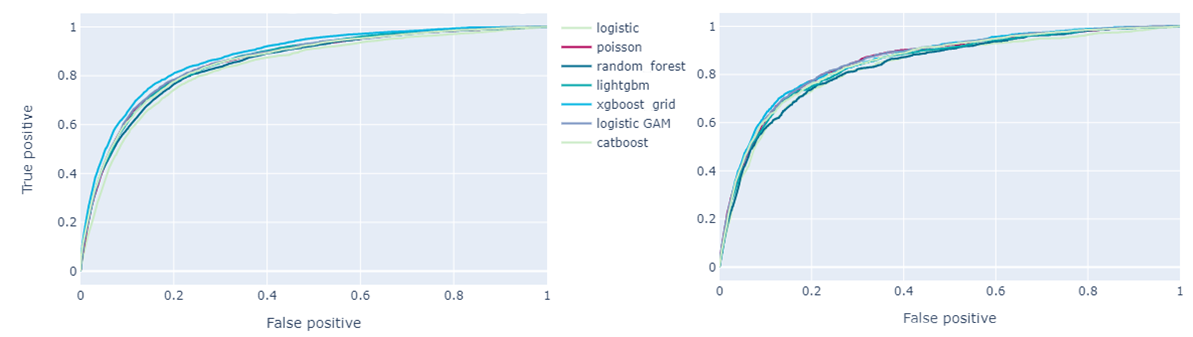}
    \caption{ROC Curve for the train (left) and test (right) datasets}
    \label{roc}
\end{figure}

The figure~\ref{roc} highlights that all our models have quite good performance levels. As expected, all of them show a better performance on the train dataset than on the test one. The performance and, consequently, the ranking of our models, are assessed based on the test data set. If the \textit{XGBoost} model seems to be the best one on the train set, it actually may be subject to over-fitting as it is no longer the best one on the test set. Based on the test dataset, the AUC is highest for the \textit{CatBoost} model.    

\begin{lstlisting}[language=Python]
from scor_survival import analysis
models = [logistic, poisson, rf, lightgbm, xgboost, gam, catboost]
analysis.plot_roc_curve(models, X, exposition, event)
\end{lstlisting}

\subsubsection{C-Index}
\textbf{Refer to section 4.2 for details on C-Index}\\ \\
For continuous models, the C-index measure could be considered as an equivalent to the AUC as it also gives insights on the risk ranking capacity of a model. Besides, it has the same order of magnitude as the AUC.\\
An important gap between the metrics obtained on the train and test datasets indicates that the model is not perfectly calibrated. The \textit{Cox-XGBoost} model seems to be slightly over-fitted, even if the C-Index on the test set is still the best one. Based on this criterion, a Cox-Net model would be preferred as it seems to be more robust.

\begin{lstlisting}[language=Python]
from scor_survival.models.continuous import model
model.ci(X, event, exposition)

from scor_survival.models.discrete import model
model.auc(X, event, exposition)
\end{lstlisting}

\begin{table}[H]
\centering
\scalebox{0.8}{
\begin{tabular}{|r|c|c|}
\hline
\textbf{Continuous Models}                       & \textit{C-Index Train} & \textit{C-Index Test}\\ \hline
\textit{\textbf{Cox}}                 & 0.86                 &  0.85                 \\
\textit{\textbf{Cox-Net}}             & 0.86                 &  0.85        \\
\textit{\textbf{Cox Tree}}            &  0.80                &  0.79   \\
\textit{\textbf{Cox XGBoost}}         & 0.89                 &  0.85                \\
\textit{\textbf{Survival Tree}}       &   0.84               &  0.83   \\
\textit{\textbf{Random Survival Forest}} &   0.85            &  0.83    \\\hline
\textbf{Discrete Models}                       & \textit{AUC Train} & \textit{AUC Test}\\ \hline
\textit{\textbf{Binomial Regression}}         & 0.83                 &  0.83                \\
\textit{\textbf{Poisson Regression}}       &   0.85               &  0.85  \\
\textit{\textbf{Random Forest}}            &  0.86                &  0.84   \\
\textit{\textbf{LightGBM}}             & 0.86                 &  0.85        \\
\textit{\textbf{XGBoost}} &   0.88            &  0.86   \\
\textit{\textbf{Logistic GAM}} &   0.86            &  0.86\\
\textit{\textbf{CatBoost}}                 & 0.86                 &  0.85                 \\
\hline
\end{tabular}}
\caption{C-Index and AUC value}
\label{ci}
\end{table}

It may be interesting to compare the predictive capacities of our model on different subgroups rather than globally. Indeed, as explained at the beginning, an insurer needs to produce consistent predictions for each category as the data used to calibrate the model is not necessarily representative of his portfolio.

\begin{lstlisting}[language=Python]
from scor_survival import analysis
models = [logistic, poisson, rf, lightgbm, xgboost, gam, catboost]
analysis.plot_pred(variable, models, X, exposition, event)
\end{lstlisting}

\begin{figure}[H]
    \hspace*{-1.4cm}
    \includegraphics[scale=0.55]{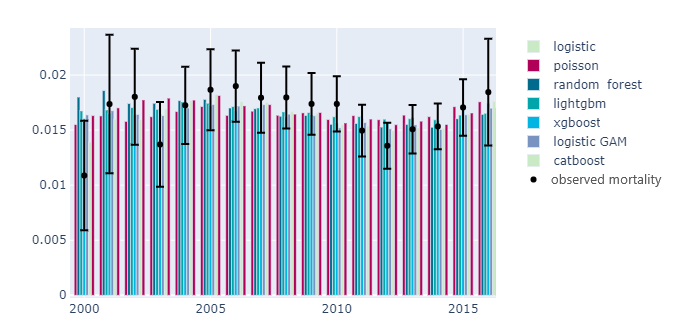}
    \includegraphics[scale=0.57]{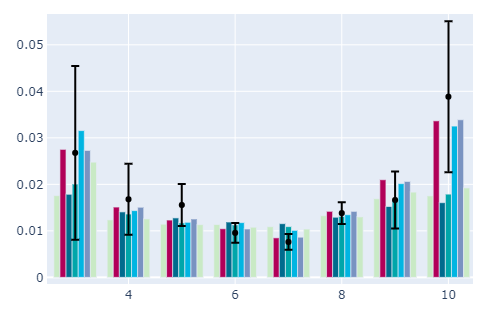}
    \caption{Mortality prediction by year of entry into the study program (left) and sleep hours (right)}
    \label{pred}
\end{figure}

In other words, we expect that the prediction is contained within the confidence interval of the observed mortality for each subgroup.\\
When considering the mortality prediction by year of entry into the study program, it seems that it is the case for all our models.\\
When focusing on mortality prediction by sleep hours, it seems that some models, mainly the random forest and the lightgbm, may underestimate the observed mortality of the dataset.

However, the selection of a model should not rely only on performance metrics, especially because model performance highly depends on the dataset on which the model is evaluated.\\
From a business angle, many other factors can influence the model selection.

\subsection{Convergence issue}

For some models, such as GAM or GLM, a gradient descent algorithm is used to estimate the parameters. Such a gradient descent algorithm can face convergence issues, especially in the situations depicted below:
\begin{itemize}
    \item If the objective function is not strictly convex, the optimization algorithm can descend into a local solution.
    \item If some variables are correlated, the problem can be intractable, and the algorithm will return an error.
    \item A large number of features (and a high degree of freedom for GAM) increases the dimension of the parameters space. The algorithm may struggle to find the optimal solution to the problem.
\end{itemize}

Within the library, the logistic GLM and GAM regressions return the logs of the internal function and raise a warning to the user in case of non convergence of the gradient descent algorithm.

\subsection{Edge effect issue}

A model is efficient where the data have been observed. When extrapolating the data with a model, the user should be very careful. In survival analysis, when the survival dataset is converted into a pseudo data table, the attained year is a numeric information. If a GAM or a GLM can regress against this variable and propose an average trend, a classification tree will only locally approximate the mortality. As such, for classification trees, the shape of the mortality prediction by attained year will flatten with increasing unobserved attained years.

\subsection{The potential of extending the goodness of fit of models: transfer learning}

Within life insurance, products differ in every country. However, depending on market maturity, experience might not always be available. One could use market benchmark data as initial input for pricing, but if it is not available, leveraging on similar market data could be useful. Nevertheless, it is not always compliant to transfer one dataset from one country to another. An alternative is then to export the knowledge accumulated on a specific dataset through the built risk model. The practice of using such a model as a starting point, called transfer learning, is a common practice in Machine Learning to face the lack of data or capacity to fit one model. \\

Usually, when it only consists in applying a model in a context that differs from its initial purpose, it could lead to bad performance. However, when models are fitted to capture actual interactions between biometric parameters and studying the influence of explanatory features on a life event (such as death), we could expect a generalization capacity. To benefit from this generalization effect, Machine Learning processes should then be controlled to not only reach high predictive performance but also to demonstrate a certain interpretability and robustness level. In practice, such control is feasible by using not only a larger panel of survival models (as presented in this document) but also transparency tooling as presented in \cite{ddint}. This enhances expert review (with doctors, underwriters,etc.) and the capacity to explain models and control the interactions models can capture.\\

Hence, this document helps in better understanding how to leverage on Machine Learning and adapt models to make them applicable to survival modeling. However, readers should be aware that applying such models cannot be reduced to only using a Python library (even if this one has been developed to limit operational risks). It is necessary to combine business knowledge and expert experience with the outcomes the models provide. 


\subsection{Business constraints}

The model calibration easiness, mainly the model sensitivity to hyper-parameters and the computation time, are important factors on the operational angle. Models may be subject to frequent updates thus could be often re-calibrated. As seen above, Random Forest or XGBoost models require expensive computing resources to be very well calibrated while regression models are generally more straightforward.\\ \\
Other important factors are the robustness and interpretability of the model. Regulation requires insurers to be able to explain every decision as well as their assessment of risk factors. In this respect, a simple Binomial model could be preferred compared to the XGBoost algorithm. However, this could change in the future thanks to the development of interpretability methods. \\
More complex models, such as Random Forest or XGBoost, have better predictive power as they can capture variables interactions and non-linear patterns. But this comes with costs regarding the calibration easiness and interpretability. An insurer has to find the right balance between model precision and business constraints.\\ \\
Based on these considerations, it seems that all our models have their strengths and weaknesses. On the NHANES dataset, the CatBoost model appears as the best compromise in terms of prediction performance, computation time, and calibration. Let's indeed remember that a CatBoost model can deal with categorical feature without prior transformation, which facilitates the process compared to all other models. 

\begin{table}[H]
\hspace*{-0.3cm}
\scalebox{0.7}{
\begin{tabular}{|l|l|l|}
\hline
 &
  \multicolumn{1}{c|}{\textbf{Strengths}} &
  \multicolumn{1}{c|}{\textbf{Weaknesses}} \\ \hline
\textit{\textbf{Binomial Regression}} &
  \begin{tabular}[c]{@{}l@{}}
  \tabitem No parameter calibration\\ 
  \tabitem Interpretability\\ 
  \tabitem Possibility to add regularization\end{tabular} &
  \begin{tabular}[c]{@{}l@{}}
  \tabitem Model only linear patterns\\
  \tabitem May not converge \end{tabular} \\ \hline
\textit{\textbf{Poisson Regression}} &
  \begin{tabular}[c]{@{}l@{}}
  \tabitem No parameter calibration\\ 
  \tabitem Interpretability\\ 
  \tabitem Possibility to add regularization\end{tabular} &
  \begin{tabular}[c]{@{}l@{}}
  \tabitem Model only linear patterns\\ 
  \tabitem May not converge\end{tabular} \\ \hline
\textit{\textbf{logistic GAM}} &
  \begin{tabular}[c]{@{}l@{}}
  \tabitem Model variable interactions and non linear patterns\\ 
  \tabitem Interpretability \\
  \tabitem Possibility to add regularization \end{tabular} &
  \begin{tabular}[c]{@{}l@{}}
  \tabitem Need to specify the interactions \\
  \tabitem May not converge \end{tabular} \\ \hline
\textit{\textbf{Random Forest}} &
  \begin{tabular}[c]{@{}l@{}}
  \tabitem Model variable interactions and non linear patterns\\ 
  \tabitem May be parallalized\\ 
  \tabitem Robust
  \end{tabular} &
  \begin{tabular}[c]{@{}l@{}}
  \tabitem Subject to overfitting\\ 
  \tabitem Sensitivity to hyperparameter\\
  \tabitem Difficulty to capture duration effect
  \end{tabular} \\ \hline
\textit{\textbf{LightGBM}} &
  \begin{tabular}[c]{@{}l@{}}
  \tabitem Model variable interaction and non linear pattern\\
  \tabitem Can adapt to large dataset with lots of features\\ 
  \tabitem Speeder gradient boosting algorithm  \end{tabular} &
  \begin{tabular}[c]{@{}l@{}}
  \tabitem Sensitivity to hyperparameters\\
  \tabitem No interpretability \end{tabular} \\
   \hline
\textit{\textbf{XGBoost}} &
  \begin{tabular}[c]{@{}l@{}}
  \tabitem Model variable interactions and non linear patterns\\ 
  \tabitem More robust compared to other gradient boosting \end{tabular} &
  \begin{tabular}[c]{@{}l@{}}
  \tabitem Subject to overfitting\\  
  \tabitem Sensitive to hyperparameters \\
  \tabitem No interpretability\end{tabular} \\ \hline
\textit{\textbf{Catboost}} &
  \begin{tabular}[c]{@{}l@{}} 
  \tabitem Model variable interactions and non linear patterns\\
  \tabitem Handle categorical variable without prior transformation\\ 
  \tabitem Less subject to overfitting compared to other gradient boosting \end{tabular} &
  \begin{tabular}[c]{@{}l@{}}
  \tabitem Sensitivity to hyperparameters\\
  \tabitem No interpretability \end{tabular} \\ \hline
\end{tabular}}
\caption{Discrete models summary table}
\end{table}

\begin{table}[H]
\hspace*{-0.3cm}
\scalebox{0.7}{
\begin{tabular}{|l|l|l|}
\hline
 &
  \multicolumn{1}{c|}{\textbf{Strengths}} &
  \multicolumn{1}{c|}{\textbf{Weaknesses}} \\ \hline
\textit{\textbf{Cox - Net}} &
  \begin{tabular}[c]{@{}l@{}}
  \tabitem No parameter calibration\\ 
  \tabitem Interpretability\\ 
  \tabitem Regularization \end{tabular} &
  \begin{tabular}[c]{@{}l@{}}
  \tabitem Model only linear patterns\\ 
  \tabitem May not converge \\
  \tabitem Rely on proportional hazard assumption \end{tabular} \\ 
  \hline
\textit{\textbf{Cox - XGBoost}} &
  \begin{tabular}[c]{@{}l@{}}
  \tabitem Model variable interactions and non linear patterns\\ 
  \tabitem Robust\\
  \tabitem Good predictive performance \end{tabular} &
  \begin{tabular}[c]{@{}l@{}}
  \tabitem Subject to overfitting\\  
  \tabitem Sensitive to hyperparameters \\
  \tabitem No interpretability\end{tabular} \\ \hline
  \textit{\textbf{Cox Tree}} &
  \begin{tabular}[c]{@{}l@{}}
  \tabitem Short computation time\\ 
  \tabitem Interpretability  \end{tabular} &
  \begin{tabular}[c]{@{}l@{}}
  \tabitem Rely on proportional hazard assumption \\
  \tabitem May not converge \\
  \tabitem Difficulty to capture duration effect \end{tabular} \\ \hline
\textit{\textbf{Survival Tree}} &
  \begin{tabular}[c]{@{}l@{}}
  \tabitem Short computation time\\
  \tabitem No underlying assumption\\ 
  \tabitem Interpretability  \end{tabular} &
  \begin{tabular}[c]{@{}l@{}}
  \tabitem Questionability of the heterogeneity measure\\
  \tabitem Difficulty to capture duration effect\end{tabular} \\
   \hline
\textit{\textbf{Random Survival Forest}} &
  \begin{tabular}[c]{@{}l@{}}
  \tabitem Model variables interactions and non linear patterns\\ 
  \tabitem May be parallalized\\ 
  \tabitem Robust 
  \end{tabular} &
  \begin{tabular}[c]{@{}l@{}}
  \tabitem Subject to overfitting\\ 
  \tabitem Sensitivity to hyperparameter\\
  \tabitem Difficulty to capture duration effect
  \end{tabular} \\ \hline
  \end{tabular}}
\caption{Time-to-event models summary table}
\end{table}

To give better insights into the comparison of the models based on this fictive portfolio, we will measure the financial impact of the different modeling strategies for an insurer. 

\section{How to derive the premium price based on estimated mortality}

\begin{figure}[H]
    \hspace*{4.5cm}
    \includegraphics[scale=0.6]{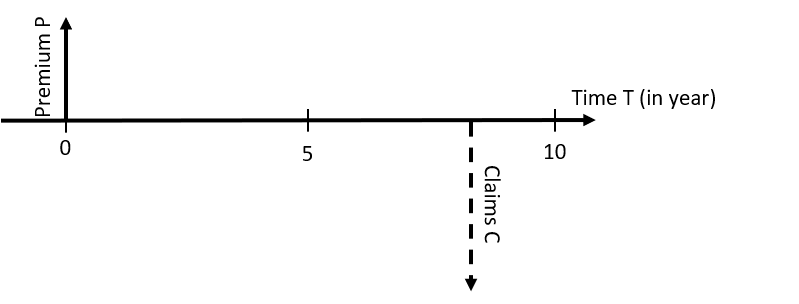}
    \caption{Life insurance contract}
    \label{cl}
\end{figure}

Survival models are used to forecast mortality rates for the insurance contract duration according to applicant characteristics. To illustrate the importance of a good mortality prediction for a life insurance company, we will consider the pricing of a policy on a competitive market. \\
Through a simulated competitive market between two insurers, we will give insights into the capacity of a model to price correctly simple products. For our purpose, we consider a simplified contract similar to mortality products that can be found in the US market (cf Figure~\ref{cl}).\\ \\
Given a premium $\pi$ paid at time 0, the underwriting date, the insured's beneficiary will earn $C$, fixed at \$ 100.000, in case of death in the following ten years. For simplicity matters, we will consider a constant interest rate $i = 1.5\%$ and we denote, $v$, the discount rate so that $v = \frac{1}{1+i}$. However, for business matters, an interest rate curve should be considered. 

The pure premium of this life insurance contract is defined as the expected claim cost given the applicant's characteristics, that is to say: \\
\begin{align*}
 \pi(X) = \mathbbm{E}[Cv^T\mid X] & = C \sum_{t=1}^{10} \mathbbm{P}(T = t\mid X)v^t \\
 & = C \sum_{t=1}^{10} S(t\mid X)h(t\mid X)\times v^t  
\end{align*}

Thus, the premium can be defined as a function of the survival curve: $\pi(X) = f_{v}(S(.|X))$. The function $f_v$ is decreasing: the higher the survival probability, the lower the insurance price. The insurer is indeed less likely to pay the claim if the death probability is low. \\
From the previous formula, it becomes obvious that a binary classification (deaths VS alive) model is not enough to price life insurance contracts as we have to be able to predict survival at several periods. Also, insurers need to precisely price insurance products given life insurance purchaser characteristics.\\

To compare the different premium estimations between the models, the figure below represents the average premium for ages from twenty to eighty years old.\\
Based on it, the age seems to impact the premium level exponentially for all models. The event of interest being death in the ten coming years, it indeed highly depends on one's age. 
Ultimately, the premium increases until almost reaching the sum insured.\\
Indeed the oldest applicants are very unlikely to survive the insurance coverage period (10 years) and thus the insurer is likely to pay the claim. With a quasi-certainty to pay the claim, the expected claim cost for the insurer - which is the pure premium - is nearing the sum insured.

\begin{figure}[H]
    \hspace*{2.9cm}
    \includegraphics[scale=0.7]{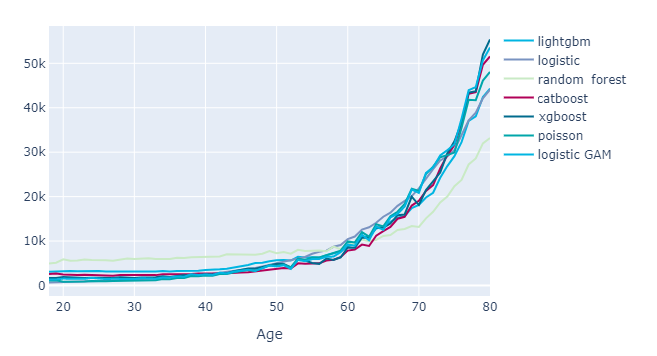}
    \caption{Discrete Model premium evaluation based on individuals' age}
    \label{premium}
\end{figure}

{\Large\textbf{Survival analysis theory}}

Two main modeling strategies exist to take censoring into account: fitting specific models to raw survival data or modifying the data structure to be able to apply standard models.\\
This chapter will introduce the survival analysis theory: the data challenges and the modeling strategies, on which Machine Learning modeling relies. 

\section{Survival Data}
\label{sc:continuous}

Before introducing the different models, the structure of survival data should be introduced. \newline
The whole theory relies on two random variables: $T$ the time to event and $C$ the censoring time. They are assumed to be independent. The censoring time is a random variable modeling the observation period of an individual, that is to say, the time between the start of the observation of the individual and the time of withdrawal or loss of tracking. The time to event is a random variable modeling the observation period between the start of observation and the studied event occurrence, for instance, the death. \newline
However, in practice, the information available in the datasets is the stopping time of observation (because of death or censoring) and an indicator of whether the observation is censored. That is to say: 

\begin{equation}
\left\{ \begin{aligned}
Y & = \min(T,C) \\ 
\delta & =  \mathbbm{1}_{\left\{T \leq C\right\}} \\
\end{aligned} \right. 
\end{equation}

Considering \textit{Age} and \textit{Gender} as explanatory variables at $t=0$, a survival data table can be built as in Table~\ref{surData} for the four previous observed individuals in Figure~\ref{ignor}.

\begin{table}[H]
\centering
\scalebox{0.9}{
\begin{tabular}{|c|cc|cc|}
\hline
\textit{\textbf{}} & \textit{\textbf{Age}} & \textit{\textbf{Gender}} & \textit{Y} & \textit{$\delta$} \\ \hline
\textit{\textbf{$S_{1}$}} & \textit{40} & \textit{Female} & \textit{7.1} & \textit{1} \\
\textit{\textbf{$S_{2}$}} & \textit{30} & \textit{Male} & \textit{4.9} & \textit{0} \\
\textit{\textbf{$S_{3}$}} & \textit{52} & \textit{Male}  & \textit{3.4} & \textit{0} \\
\textit{\textbf{$S_{4}$}} & \textit{60} & \textit{Female} & \textit{3} & \textit{1} \\ \hline
\end{tabular}}
\caption{Example of survival data table}
\label{surData}
\end{table}

The columns are divided into three categories, the first two columns representing the risk factors, Y being the end of the follow-up period observed, and $\delta$ an indicator of death.\newline 
From this example, we can deduce that the first individual, who is a forty-year-old woman dies after 7 years whereas the second individual is censored after 4.9 years.

\section{Quantity of interest}

The theory focuses on two functions as the quantity of interest to estimate: the \textbf{survival function $S$} and the \textbf{hazard function $h$}. Having an estimation of one of them allows to fully model the survival of an individual. \\

\begin{figure}[H]
    \centering
    \includegraphics[width=0.6\textwidth]{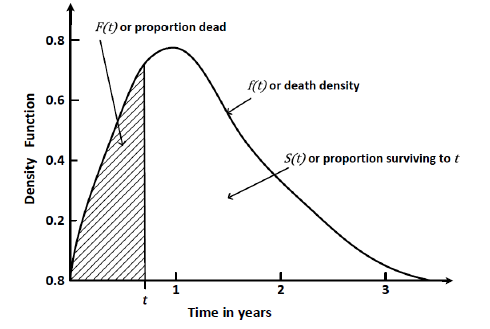}
    \caption{Relationships between $f(t)$, $F(t)$ and $S(t)$ (source \cite{figure})}
    \label{fig:survival}
\end{figure}

The \textbf{survival function $S$} represents the probability that the time to event is not earlier than a specific time $t$:
\begin{equation}
    S(t) = Pr(T \geqslant t)
\end{equation}
The survival function decreases from 1 to 0. The meaning of a probability equal to 1 at the starting time is that 100\% of the observed subjects are alive when the study starts: none of the events of interest have occurred. From this quantity, we can define the \textit{cumulative death distribution function} $F(t) = 1 - S(t)$ and the \textit{density function} $f(t) = \frac{dF(t)}{dt} = \frac{-dS(t)}{dt}$ for continuous cases and $f(t) = \frac{[F(t + \Delta t) - F(t)]}{\Delta t}$ for discrete cases.
The relationships between these functions is shown in Figure \ref{fig:survival}.

The second quantity of interest is the \textbf{hazard function $h$}. It indicates the occurrence rate of an event at time $t$, given that no event occurred before. Formally, the hazard rate function is defined as:
\begin{equation}
\begin{aligned}
h(t) & = \lim_{\Delta t \to 0} \frac{Pr(t \leqslant T \leqslant t + \Delta t | T \geqslant t)}{\Delta t} \\
& = \lim_{\Delta t \to 0} \frac{F(t + \Delta t) - F(t)}{\Delta t S(t)} \\ 
& = - \frac{\mathrm{d}\, \log S(t)}{\mathrm{d}\,t}
\end{aligned}
\end{equation}

From this equation we can easily derive that $$S(t) = exp(-\int_{0}^{t} h(s)\mathrm{d}s) = exp(-H(t))$$
where $H(t) = -\int_{0}^{t} h(s)\mathrm{d}s $ is called \textit{the cumulative hazard function}.\\
Using the same notation as before, we can define a likelihood function taking into account censoring:
\begin{equation}
    L = \prod_{i} P(T=t_{i})^{\delta_{i}}P(T>t_{i})^{1-\delta_{i}} = \prod_{i} h(t_{i})^{\delta_{i}}S(t_{i})
\label{eq:likelihood}
\end{equation}

The intuition of the function comes from the contribution to the likelihood function between a censored and a full-observed individual: 
\begin{itemize}
    \item If an individual dies at time $t_{i}$, its contribution to the likelihood function is indeed the density that can be written as $S(t_{i})h(t_{i})$. 
    \item If the individual is still alive at $t_{i}$, all we know is that the lifetime exceeds $t_{i}$, which means that the contribution to the likelihood function is $S(t_{i})$.
\end{itemize}

\section{Non-parametric survival model}

Non-parametric models present many advantages. Contrary to parametric ones, for which convergence issues may appear during the optimization steps due to correlation or non convexity, non-parametric models will always produce the desired estimator. Then, due to the reliance on fewer assumptions, non-parametric methods are quite robust and may be applied even in situations where little is known about the application in question. Finally they are less time and memory consuming, which presents a real goal when dealing with large amounts of data.

\subsection{Kaplan-Meier estimator}
\label{sc:KM}

When we have no censored observations in the data, the empirical survival function is estimated by:
\begin{equation}
    \hat{S}(t) = \frac{1}{n} \sum_{i=1}^{n} \mathbbm{1}_{t_{i} \leq t}
\end{equation}
This estimation is no longer viable in presence of censoring as we do not observe the death time $t_{i}$ but the end of observation time $y_{i}$. Thus \citet{KM} extended the non-parametric estimation to censored data (see Appendix~\ref{ch:KM} for details):
\begin{equation}
\hat{S}(t)=\prod \limits _{t_{i}\leq t}(1- \frac{d_{i}}{N_{i}})^{\delta_{i}} 
\end{equation}

Based on the four individuals of Figure~\ref{ignor}, the survival probability at time 5 is equal to $\hat{S}(5) = (1-\frac{1}{4})^{1} \times (1-\frac{0}{3})^{0} \times (1-\frac{0}{2})^{0} = \frac{3}{4}$ \\
Computing this value for each period enables us to obtain a step function for the survival function where the jumps are observed at the empirical observed death times.

\begin{figure}[H]
    \centering
    \includegraphics[scale=0.5]{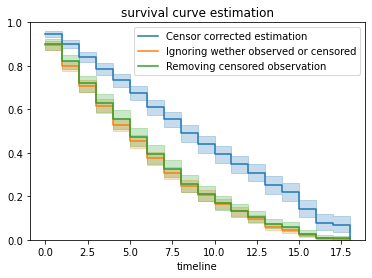}
    \caption{\textbf{Impact of ignoring censoring in life duration study}}
    \label{cens}
\end{figure}

As introduced before, ignoring censoring leads to an underestimation of the life duration. Figure~\ref{cens} highlights this underestimation. Three 'Kaplan-Meier' survival curves are plotted on different datasets: the real one, the one relying only on fully observed individuals, and the one for which censored and dead individuals are not distinguished. As the two last curves are below the real one, it means that at each time the survival probability is lower and thus that risks have been overestimated. \\ \\
Kaplan-Meier estimator is the most widely used because of its simplicity to compute. It is implemented in many survival libraries and packages of statistical and mathematical software. Besides, this estimator doesn't rely on any assumption and can thus easily be used as a reference model or to test hypothesis. \\
It is effective to get the survival curve of the global population. However, the precision of the estimation relies on the number of observations. If we want to take into account individuals' characteristics, we need to recompute the estimator for each chosen subset, which reduces the number of observations and thus the accuracy. \\
On the business side, it is indeed important to have a good prediction among different subgroups rather than on the global level. The insurer portfolio may indeed have an over-representation of some individuals compared to the population used to build the model, keeping in mind that the insured population has lower mortality compared to the global population.

\subsection{Nelson-Aalen}

Instead of estimating the survival function, another method has been developed by~\citet{NA} and~\citet{NA2} to estimate the cumulative hazard function. 
Using the previous notations, it is defined as:
$$\hat{H}(t) = \sum_{t_i \leq t} \frac{d_i}{N_i}\delta_i$$
Based on the four individuals of Figure~\ref{ignor}, the cumulative hazard function at time 5 is equal to $\hat{H}(5) = \frac{1}{4}$ \\
To get an estimator of the survival function, one only has to plug-in the cumulative hazard estimator into the formula $S(t) = e^{-H(t)}$, in the example, $\hat{S}(5) = exp^{-\frac{1}{4}} = 0.77$
$$\hat{S}(t) = e^{-\hat{H}(t)} = \prod_{t_i \leq t} (e^{-\frac{d_i}{N_i}})^{\delta_i} \approx \prod_{t_i \leq t} (1-\frac{d_i}{N_i})^{\delta_i}$$
If the number of deaths is small compared to the number of people at risk at each time, the Nelson-Aalen plug-in survival function can be approximated by the Kaplan-Meier estimator. The two estimators are thus numerically close, but they have different properties, which implies different confidence intervals or median times (see \cite{German} for details on estimator properties).

\section{Cox Model}

Cox's model~\cite{Cox} allows to take the effect of covariates into account and to measure their impacts through the estimation of the hazard function. It then possible becomes to rank people's risk according to their characteristics. This model may be considered as a regression model for survival data, which is a particular case of the proportional hazard models.
Such models are expressed as a multiplicative effect of the covariates on the hazard function through the expression:
\begin{equation}
    h(t \ | \  X) = h_{0}(t) \times g(\beta' X)
\end{equation}

\begin{itemize}
	\item[$X$] the vector of covariates which must be time-independent
    \item[$g$] a positive function
    \item[$\beta$] the parameter of interest
    \item[$h_{0}$] the baseline hazard function for individuals with X=0.
\end{itemize}

Given two individuals $A$ and $B$ with covariates $X_{A}$ and $X_{B}$ respectively, the ratio of their hazard functions is assumed to be unchanged over time. That is the reason why such models are said to be proportional hazard models.\newline
For the particular case of the Cox model, the function $g$ is an exponential function as the Cox model is defined by  $ h(t \ | \  X) = h_{0}(t) \times exp(\beta' X)$. 
The main interest of the model is the possibility to rank people on their risk level without computing the survival function. The relative risk is introduced to this end: 
\begin{equation}
RR = \frac{h(t|X_{A})}{h(t|X_{B})} = exp(\beta(X_{A}-X_{B}))    
\end{equation}
The estimation can be divided into two steps. Depending on the purpose of the study, one may stop at the first one.

\subsubsection{Estimation of the risk parameter $\beta$}
We compute the estimator $\hat{\beta}$ by maximizing the partial likelihood function defined by Cox (cf Appendix~\ref{ch:Cox}) :
\begin{equation}
    L(\beta) = \prod _ {i=1} ^ m \frac{e^{X'_{j_{(i)}} \beta}}{\sum _ {j \in R_i} e^{X'_{j} \beta}}
    \label{coxLL}
\end{equation}
\begin{itemize}
\item[] $m$ the total of uncensored individuals
\item[] $j_{(i)}$ the individuals who died at time $t_{(i)}$
\item[ ] $t_{(i)},...,t_{(m)}$ the ordered time of observed death events
\item[] $R_{i}$ the risk set, which is a set of indices of the subjects that were still alive just before $t_{(i)} : R_{i} = \{j : t_{j} \leq t_{(i)}\}$ 
\end{itemize}

Based on the same four observed individuals in Figure~\ref{ignor} and considering \textit{Female} as the baseline category,

\begin{equation}
    L(\beta) = \frac{e^{60\beta_{2}}}{e^{60\beta_{2}} + e^{\beta_{1} + 52\beta_{2}} + e^{\beta_{1} + 30\beta_{2}} + e^{40\beta_{2}}} \times \frac{e^{40\beta_{2}}}{e^{40\beta_{2}}}
    \label{coxEx}
\end{equation}

The maximum likelihood estimators are obtained for $\hat{\beta_{1}} = -10.17$ and $\hat{\beta_{2}} = 0.84$. \\
When one is only interested in comparing the survival curve to classify individuals according to their survival probabilities, only the estimation of the risk parameter $\beta$ is needed. The baseline hazard $h_{0}(t)$ does not have any effect on the relative risk. Indeed, in this case we can deduce that a \textit{Woman} will have a higher mortality compared to a \textit{Man}, and that the mortality increases with the \textit{Age}. 

\subsubsection{Estimation of the baseline function}
If one needs the survival function for every individual, for premium computation for instance, the hazard baseline $h_{0}(t)$ is needed in addition of the $\beta$ parameters. The survival function can be computed as follows:
\begin{equation}
    \hat{S}(t) = exp(-H_0(t) exp(X\hat{\beta})) = S_0(t)^{exp(X\hat{\beta})}
\end{equation}
where $H_0(t)$ is the cumulative baseline hazard function, and $S_0(t) = exp(-H_0(t))$ represents the baseline survival function. Breslow’s estimator is the most widely used method to estimate $\hat{S}_0(t) = exp(-\hat{H}_0(t))$ where 
$$\hat{H}_0(t) = \sum_{t_{i}<t} \hat{h}_{0}(t_{i}) \text{ with } \hat{h}_{0}(t_{i}) = \frac{1}{\sum_{j \in R_{i}} e^{X_{j}\beta}} \text{ if } t_{i} \text{ is a time event, 0 otherwise}$$
Based on our example, with the observed individuals of Figure~\ref{ignor}, we can deduce:
$\hat{S}_{0}(5) = 1$ and thus for a 43 year-old-man $\hat{S}(5)=1^{exp(-10.17+43 \times 0.84)}=0.99$ and for a 43 year-old-woman $\hat{S}(5)=1^{exp(-10.17)}=0.56$. The average mortality between the two individuals is $0.77$, which is indeed the global mortality found with the Kaplan-Meier estimator.

\section{Exposures: an actuarial approach}
\label{sc:discrete}

Another method, widely used in actuarial science, specifically to build mortality tables, consists in discretizing the data into small time intervals. The discretization enables us to apply traditional methodologies to predict mortality, as it removes the lack of information due to censoring thanks to the exposure to risk.\\ \\
Withdrawal from the study of the censored subjects introduces bias if we compute traditional estimators. The mortality rate, $q_{j}$, within a time interval $[\tau_{j},\tau_{j}+1]$, denoted interval $j$, can no longer be estimated with the ratio of the deaths, $d_{j}$, on the number of alive subjects at the beginning of the interval, $l_{j}$. The quantity $\frac{d_{j}}{l_{j}}$ is indeed an inaccurate estimation as deaths that occur after withdrawal will not be known. Therefore, withdrawing life is only exposed to the risk of death.\\
To compensate for the withdrawal, the number of alive subjects, $l_{j}$, is replaced by the number of subjects exposed to risk. Depending on the hypothesis made on mortality, several types of exposure can be considered. \\
Three exposures are considered by actuaries: \textit{Distributed exposure}, \textit{Initial exposure} and \textit{Central exposure}. However, we will only focus on the last two as the distributed exposure method relying on uniform distribution of deaths is not currently a widely-used one (see \cite{SoA} for details).
 
\subsection{Initial Exposure and Balducci hypothesis}

We denote \textbf{initial exposure} the quantity $EI_{j}$, which represents the global amount of time each life was exposed to the risk of death during the interval $j$. As the exposure is based on the lives at the start of the interval the exposure can be referred to as initial. \\
$EI_{j}$ is the aggregation of the following individual exposure, $ei_{j}$: 

\begin{itemize}
    \item[$\bullet$] Alive at the start and the end of the interval are assigned 1
    \item[$\bullet$] Deaths during the time interval are assigned 1
    \item[$\bullet$] Censored are assigned the fraction of the interval during which they were observed 
\end{itemize}

Formally, if we respectively denote $c_{i,j}$ and $t_{i,j}$ the censoring and death time of the individual i in interval j, $w_{j}$ the number of withdrawals and $l_{j}$ the number of alive subjects, the \textbf{initial exposure} is expressed as (see Appendix~\ref{ch:EI} for intuition on the formula):
\begin{align*}
 EI_{j} & = \sum_{i}^{l_{j}} \mathbbm{1}_{\left\{t_{i,j} > 1\right\}}\times\mathbbm{1}_{\left\{c_{i,j} > 1\right\}}  +
     \mathbbm{1}_{\left\{t_{i,j} < 1\right\}} +  c_{i,j}\mathbbm{1}_{\left\{c_{i,j} < 1\right\}} \\
 & =  \sum_{i}^{l_{j}} 1 - \mathbbm{1}_{\left\{c_{i,j} < 1\right\}} +  c_{i,j}\mathbbm{1}_{\left\{c_{i,j} < 1\right\}} \\ 
 & = l_{j} - w_{j} + \sum_{i=1}^{w_{j}} c_{i,j}
\end{align*}
To understand the idea behind the quantity, let's define the two following notations for the rate of mortality:
\begin{itemize}
    \item $q_{j} = \mathbbm{P}(T \leq \tau_{j}+1 | T > \tau_{j})$ in interval $j$
    \item ${}_{c_{i,j}}q_{j} = \mathbbm{P}(T \leq \tau_{j} + c_{i,j} | T > \tau_{j})$ for the one in the interval $[\tau_{j},c_{i,j}]$
\end{itemize}
The number of deaths can be expressed as the sum of deaths observed within the interval and deaths expected for the censored subjects. Formally:
\begin{equation}
    d_{j} = (l_{j} - w_{j})q_{j} + \sum_{i=1}^{w_{j}} {}_{c_{i,j}}q_{j} = l_{j}q_{j} - \sum_{i=1}^{w_{j}}{}_{1-c_{i,j}}q_{j + c_{i,j}}
\end{equation}

The \textbf{Balducci hypothesis} supposes that mortality rates decrease over the interval and are defined as: \\
\begin{align*}
    {}_{1-c_{i,j}}q_{j+c_{i,j}} & = P(T_{i} \leq \tau_{j}+1 | T_{i} > \tau_{j} + c_{i,j}) \\
    & = (1 - c_{i,j}) P(T_{i} \leq \tau_{j}+1 | T_{i} > \tau_{j}) = (1 - c_{i,j})q_{j}
\end{align*}

Injecting it in the previous equation gives:
$$d_{j} = l_{j}q_{j} - q_{j}\sum_{i}^{M}(1-c_{i,j})$$
Solving the formula for $q_{j}$:
$$\hat{q}_{j} = \frac{d_{j}}{l_{j} - \sum_{i=1}^{w_{j}}(1-c_{i,j})} = \frac{d_{j}}{l_{j} - w_{j} + \sum_{i=1}^{w_{j}}c_{i,j}} = \frac{d_{j}}{EI_{j}}$$
We finally get the rate of mortality estimator corrected for censoring with the previous definition of \textbf{initial exposure} as expected. This approach relies on the \textit{Balducci assumtion}, which generally does not fit for mortality, as mortality  rates increase with time. However, withdrawals are usually small compared to the population, which allows to ignore these errors.

\subsection{Central Exposure and constant hazard function assumption}

Depending on the mortality observed within a dataset, one may prefer to use another assumption: the constant hazard function over each time interval. In this case, another exposure should be used.\\
The \textbf{central exposure}, $EC_{j}$ is the time individuals are observed within the interval. The difference with the \textbf{initial exposure} is that only individuals who survived during the whole time interval are assigned 1. \\ 
The \textit{constant hazard function} assumption implies that the hazard is constant over each time interval.\\
For $e\;\in\;[0,1]$, we denote $h_{j}$ the hazard rate over the interval $[\tau_{j}, \tau_{j}+1]$: 
\begin{equation}
h(\tau_{j}+e) = h_{j}    
\end{equation}

As long as we consider small enough time intervals, this hypothesis is acceptable. \\
When $h_{j}$ is known for each $j$, the survival function is easy to compute: \\
\begin{equation}
S(\tau_{j}+e) = \exp( -\int_{0}^{\tau_{j}+e} h(s)ds) = \exp( -\sum_{s=1}^{j-1} h_{s} + eh_{j})    
\end{equation}
The goal is then to estimate each $h_{j}$.\\
Let $ec_{i,j}$ be the \textbf{individual central exposure}, which corresponds to the time one is observed within an interval. In addition, $\delta_{i,j}$ is a death indicator in $[\tau_{j},\tau_{j+1}]$ (1 if death is observed, 0 otherwise). The likelihood can then be written as:
\begin{equation}
L = \prod_{i} S(\tau_{j}+ec_{i,j})h(\tau_{j}+ec_{i,j})^{\delta_{i,j}}    
\end{equation}
Using the constant hazard function assumption and considering the logarithm of the likelihood, we get: 
\begin{equation}
    \log(L) = \sum_{i} [ec_{i,j}h_j + \delta_{i,j}\log(h_j) - \sum_{s=1}^{j-1} h_{s}]
\end{equation}
The maximum likelihood estimator $\hat{h}_{j}$ so that $\frac{d}{dh_j}log (L) = 0$ is then the ratio of the number of deaths observed within the interval divided by the exposure:
\begin{equation}
\hat{h}_{j} = \frac{\sum_{i}\delta_{i,j}}{\sum_{i}ec_{i,j}} = \frac{d_j}{EC_j}
\end{equation}
By definition, we can write $\hat{q}_{j} = 1 - exp(-\hat{h}_{j} )$. 

As for the initial exposure, the central exposure is interesting because it can be expressed through a closed formula. However, it also relies on a death distribution assumption, which is generally not verified in practice.

\subsection{Modeling using exposure}
The main advantage of discretization is that it allows to consider classical modeling approaches, by predicting the number of deaths for each time interval. In practice, we will model the random variable $d_{j}$ describing the number of deaths using the exposure as weights or offset. Exposures are easy to compute and take into account censoring. However, this approach can generate a high number of lines in the dataset as we need to create a \textit{pseudo data table}, slowing down the computation. 

\subsubsection{Pseudo data tables}
\label{sec:pseudo}

Models can be applied to pseudo data tables, which are alternative data structures in survival analysis modeling. Often, in the actuarial field, the information of a portfolio is directly presented in pseudo data tables. If not, we can easily transform traditional survival data tables into pseudo data tables.\\
In practice, we have to generate for each individual as many rows as time intervals and for each of them compute individual exposure. The size of the intervals is fixed in advance: months, quarters, years, etc. The size choice depends on how granular and accurate the output is needed.\\

The last time interval includes the time of death or censoring, which means that $\delta$ is always equal to $0$ except for the last time interval. Regarding the exposure, it is always equal to 1 except for the last time interval, where it represents the time observed in any case when we compute the \textbf{central exposure} or only in case of censoring when we compute the \textbf{initial exposure}.\\ 
Finally, we build the dataset for every individual in the study as illustrated for the four following observations in Table~\ref{ExpData}. 

\begin{figure}[H]
    \centering
    \includegraphics[scale=0.4]{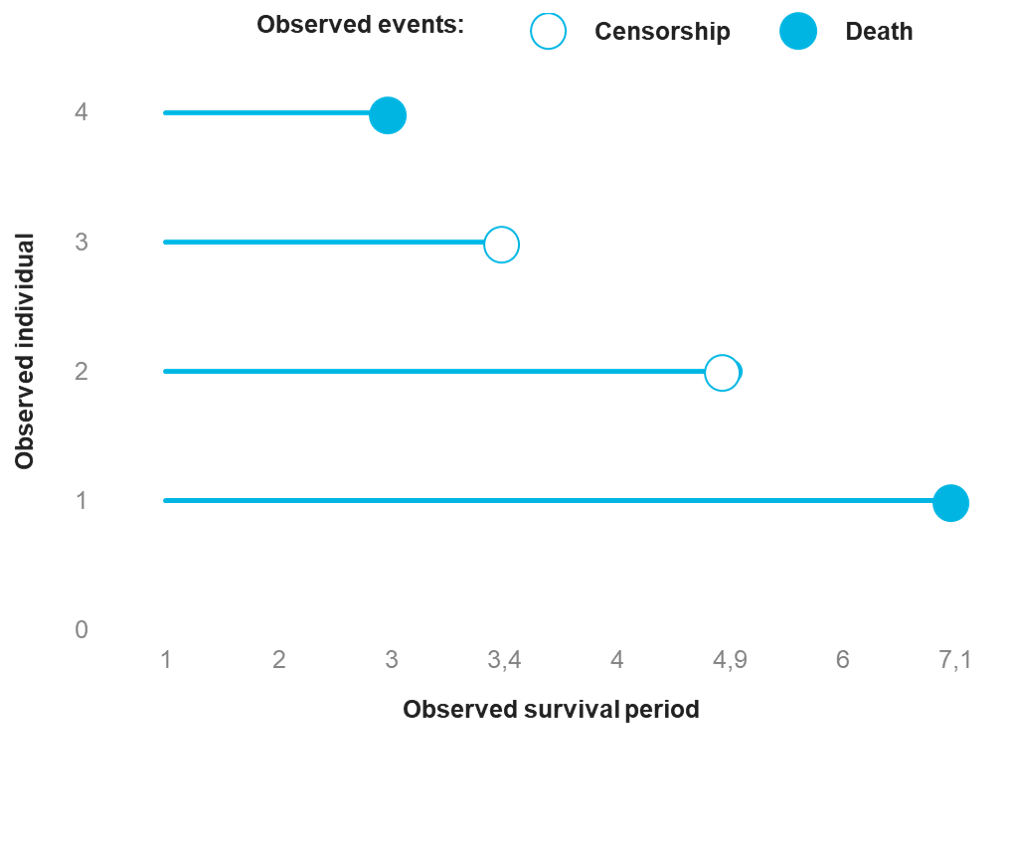}
    \label{fig:my_label}
\end{figure}

It is worth noticing that this approach allows to consider covariates that vary with the time interval. It is the case for time-varying covariates such as smoking habits. It is one advantage of this method in opposition to previous approaches considering only information at the start of observation. \\
The time interval $j$ is added to the feature variables. That is to say, that the same individual is seen as two different ones depending on the time interval $j$ considered.\\

\begin{table}[H]
\centering
\scalebox{0.85}{
\begin{tabular}{|c|cc|cccc|}
\hline
\textit{\textbf{}}  & \textit{\textbf{Age}} & \textit{\textbf{Gender}} & \textit{\textbf{I}} & \textit{\textbf{ec}} & \textit{\textbf{ei}} & \textit{\textbf{$\delta$}} \\ \hline
\textit{\textbf{$S_{10}$}} & \textit{40} & \textit{Female} & \textit{0} & \textit{1} & \textit{1} & \textit{0} \\
\textit{\textbf{$S_{11}$}} & \textit{41} & \textit{Female} & \textit{1} & \textit{1} & \textit{1} & \textit{0} \\
\textit{\textbf{$S_{12}$}} & \textit{42} & \textit{Female} & \textit{2} & \textit{1} & \textit{1} & \textit{0} \\
\textit{\textbf{$S_{13}$}} & \textit{43} & \textit{Female} & \textit{3} & \textit{1} & \textit{1} & \textit{0} \\
\textit{\textbf{$S_{14}$}} & \textit{44} & \textit{Female} & \textit{4} & \textit{1} & \textit{1} & \textit{0} \\
\textit{\textbf{$S_{15}$}} & \textit{45} & \textit{Female} & \textit{5} & \textit{1} & \textit{1} & \textit{0} \\
\textit{\textbf{$S_{16}$}} & \textit{46} & \textit{Female} & \textit{6} & \textit{1} & \textit{1} & \textit{0} \\
\textit{\textbf{$S_{17}$}} & \textit{47} & \textit{Female} & \textit{7} & \textit{0.1} & \textit{1} & \textit{1} \\\hline

\textit{\textbf{$S_{20}$}} & \textit{30} & \textit{Male} & \textit{0} & \textit{1} & \textit{1} & \textit{0} \\
\textit{\textbf{$S_{21}$}} & \textit{31} & \textit{Male} & \textit{1} & \textit{1} & \textit{1} & \textit{0} \\
\textit{\textbf{$S_{22}$}} & \textit{32} & \textit{Male} & \textit{2} & \textit{1} & \textit{1} & \textit{0} \\
\textit{\textbf{$S_{23}$}} & \textit{33} & \textit{Male} & \textit{3} & \textit{1} & \textit{1} & \textit{0}\\
\textit{\textbf{$S_{24}$}} & \textit{34} & \textit{Male} & \textit{4} & \textit{0.9} & \textit{0.9} & \textit{0}\\ \hline 

\textit{\textbf{$S_{30}$}} & \textit{52} & \textit{Male} & \textit{0} & \textit{1} & \textit{1} & \textit{0} \\
\textit{\textbf{$S_{31}$}} & \textit{53} & \textit{Male} & \textit{1} & \textit{1} & \textit{1} & \textit{0} \\
\textit{\textbf{$S_{32}$}} & \textit{55} & \textit{Male} & \textit{2} & \textit{1} & \textit{1} & \textit{0}\\
\textit{\textbf{$S_{33}$}} & \textit{55} & \textit{Male} & \textit{3} & \textit{0.4} & \textit{0.4} & \textit{0} \\ \hline 

\textit{\textbf{$S_{40}$}} & \textit{60} & \textit{Female} & \textit{0} & \textit{1} & \textit{1} & \textit{0} \\
\textit{\textbf{$S_{41}$}} & \textit{61} & \textit{Female} & \textit{1} & \textit{1} & \textit{1} & \textit{0} \\
\textit{\textbf{$S_{42}$}} & \textit{62} & \textit{Female} & \textit{2} & \textit{1} & \textit{1} & \textit{0}\\
\textit{\textbf{$S_{43}$}} & \textit{63} & \textit{Female} & \textit{3} & \textit{0} & \textit{1} & \textit{1} \\ \hline

\end{tabular}}
\caption{Example of survival data table exposure transformation}
\label{ExpData}
\end{table}
{\Large\textbf{Performance computation and model validation}}

Due to the very nature of survival data, classic metrics such as the Area Under the ROC curve (AUC) or the Mean Squared Error (MSE) might not be adapted to measure the model performances. Also, censoring prevents us from directly applying usual metrics.\\
Statisticians proposed several metrics to deal with survival data along with estimators when the observed survival is censored. In this section, we describe some of these metrics.

\section{Standardized Mortality Ratio}
\label{sec:SMR}

One of the most common and widely used metrics is the Standardized Mortality Ratio (SMR). Also known as Actual to Expected ratio in the actuarial field, the SMR can be used to measure the prediction accuracy of a model. It is defined as the ratio between the number of observed deaths and the number of deaths predicted by the model.

\begin{equation}
    \mathrm{SMR}=\frac{\sum_i \delta_i}{\sum_i pred_i}
\end{equation}

with $\delta_i=1$ if we observe the death of individual $i$ and $\delta_i=0$ otherwise. The total number of dexpected eaths is obtained by summing $pred_i$ defined as the model predicted probability to observe the death of individual $i$.

A SMR close to 1 indicates that the model fits the observations well. Different values indicate that the model may have a bias. A SMR lower than 1 shows that the model overestimates the mortality, while a SMR higher than 1 indicates that the model underestimates the mortality. 

Censoring must be taken into account when estimating the death probabilities $pred_i$. Indeed, it is unlikely to observe the death of an individual that has left the study after only a few days, while for an individual observed several years the probability should be higher.\\
Using the law of large numbers, the probability to observe a death within the study period can be approximated by the sum of the observed death divided by the number of observations. But we must add the number of non-observed deaths because of censoring. Let's start with the following equation:

\begin{equation}
    \mathrm{P}(T\leq\tau)\simeq \frac{1}{N}\sum_{i=1}^N \delta_i + \frac{1}{N}\sum_{i=1}^N (1-\delta_i)\mathrm{P}(T\leq\tau\vert\,T>t_i)
\end{equation}
with $t_i$ the observation period for person $i$, $\tau$ the maximum observation period and $T$ the random variable modeling the survival time. After a few simplifications we end up with the following equality:
\begin{equation}
    \sum_i \delta_i \simeq \sum_i (1-S(\tau))-(1-\delta_i)\left(1-\frac{S(\tau)}{S(t_i)}\right)
\end{equation}
We recall the survival function definition: $S(t)=\mathrm{P}(T > t)$. Considering $\hat{S}(t\vert\,X_i)$ the survival probability predicted by the model, we can define $pred_i$ as follows: 
\begin{equation}
pred_i=1-\hat{S}(\tau\vert\,X_i)-(1-\delta_i)\left(1-\frac{\hat{S}(\tau\vert\,X_i)}{\hat{S}(t_i\vert\,X_i)}\right)    
\end{equation}

\section{Concordance Index}
\label{sc:CI}

The Concordance Index, also called C-Index, has been introduced by \citet{Harrell}. Mainly used to measure the relevancy of the bio-marker for the survival estimation, this metric is also used to assess the predictive performance of survival models.

This metric allows us to measure the model's ability to rank individuals according to their survival. This metric is very relevant when the main goal of the model is to classify individuals according to their mortality risk, i.e. rank individuals from the ones with the lowest mortality to the ones with the highest mortality. This metric measures the classification ability of the model, but it does not measure fit quality. Thus the potential bias of a model would not be detected by this metric, which is thus complementary to the SMR.

The C-Index is defined as a conditional probability: the model survival predictions $(M_i,M_j)$ of the two individuals $i$ and $j$ are ranked in the same way as their respective survival observations $(T_i,T_j)$.
\begin{equation}
    \mathrm{C-Index} = \mathrm{P}\left(M_i<M_j\vert\, T_i<T_j\right)
\end{equation}
As the survival model predicts $M_i$, one could consider the predicted life span $M_i=\mathrm{E}[T\,\vert\,X_i]$ or the survival up to the end of the study period probability $M_i=\hat{S}(\tau\vert\,X_i)$. Note that, the classic definition is $\mathrm{C-Index} = \mathrm{P}\left(M_i>M_j\vert\, T_i<T_j\right)$ as the model score $M_i$ is considered. As the higher the score, the higher mortality, the score and survival observation of the pairs must be ranked in opposite orders. However, in this study, we found more convenient ways to consider the expected survival period.

A C-Index close to 1 indicates good performance of the model while a C-Index close to 0.5 indicates poor performance.

The C-Index can be estimated only on the pairs of observations $(i,j)$ that are comparable. Indeed, due to censorship, some pairs are not comparable. In Figure~\ref{figure_censor_illustration} we provide an illustration of this issue. In this example, in pair $(2,3)$ both survival observations are censored, therefore we are not able to tell which of the two individuals has the highest survival period. Similarly, in pair $(1,2)$ we cannot tell who has survived longer, as we are losing track of individual $2$ after year 5.

\begin{center}
\includegraphics[scale=1]{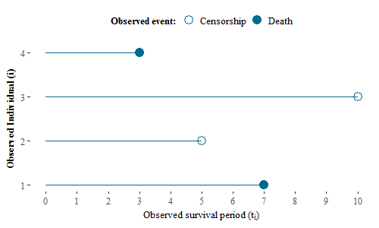}
\captionof{figure}{Censorship illustration, only pairs $(1,3), (1,4), (2,4)$ and $(3,4)$ are comparable.}
\label{figure_censor_illustration}
\end{center}

Let $\Omega$ denote the ensemble of comparable pairs $(i,j)$ where $T_i<T_j$, then we can estimate the C-Index as follows:
\begin{equation}
    \mathrm{C-Index} = \frac{1}{\mathrm{Card}(\Omega)}\sum_{(i,j)\in \Omega}  \mathds{1}_{\left\{ M_i<M_j \right\} }
\end{equation}

However, this estimator depends on the study-specific censoring distribution. While this should have limited impact when using the C-Index to compare model performances in a study, this prevents an accurate comparison of the C-Index from one study to the other. In order to have a better estimation of the C-Index, \citet{Uno} proposed an estimator based on the ICPW\footnote{Inverse Censorship Probability Weighting} approach. They proposed a free of censoring distribution estimator as follows: 
\begin{equation}
    \mathrm{C-Index} = \frac{\sum_i\sum_j \Delta_j G(t_j)^{-2} 
\mathds{1}_{\left\{t_i<t_j
\right\}} \mathds{1}_{\left\{M_i<M_j \right\}} }{ \sum_i\sum_j \Delta_j G(t_j)^{-2} \mathds{1}_{\left\{t_i<t_j \right\}} }
\end{equation}
where $G(t)$ denotes the probability of not having censoring up to time $t$, and $\Delta_j=1$ if no censoring, 0 otherwise.

\section{Brier Score}
Initially, the Brier Score was introduced by \citet{Brier} to measure the accuracy of meteorological forecasts. Then, \citet{GRAF} proposed to use this metric in the bio-statistics field for assessing survival model performance. As the interpretation of it is quite difficult, it is mainly used for model comparison. 

The Brier Score, denoted BS, is defined as the average of squared difference between the survival probabilities and the survival observations a a given time $t$.
\begin{equation}
    \mathrm{BS}(t)=\frac{1}{N}\sum_i\left(\mathds{1}_{\left\{T_i>t\right\}}-\hat{S}(t\vert\,X_i)\right)^2
\end{equation}
with $\hat{S}(t\vert\,X_i)$ the survival probability predicted by the model.

Because of censoring the BS cannot be estimated with the previous formula. Indeed, if censoring occurred before the fixed time $t$, we cannot know if the individual has survived longer than $t$. As for the C-Index, the authors considered an ICPW approach and they proposed the following estimator:
\begin{equation}
    \mathrm{BS}(t)=\frac{1}{N}\sum_i\frac{\hat{S}(t\vert\,X_i)^2}{G(t_i)}\mathds{1}_{\left\{t_i\le t \,;\,d\delta_i=1\right\}}+
\frac{\left(1-\hat{S}(t\vert\,X_i)\right)^2}{G(t)}\mathds{1}_{\left\{t_i>t\right\}}
\end{equation} where $G(t)$ is the probability of not observing censoring up to time $t$.

Like the Mean Squared Error, the lower the BS the better. Usually, for model comparison, the Brier Skill Score, BSS, is considered. It is defined as the reduction of the BS compared to the BS obtained on a reference model. \begin{equation}
    \mathrm{BSS}(t)=1-\frac{\mathrm{BS}(t)}{\mathrm{BS}_{ref}(t)}
\end{equation}

One needs to specify the time $t$ to compute the Brier Score. Depending on the purpose, a specific time $t$ can be more relevant, for instance, we are studying the survival up to 5 years. Alternatively, the time-independent metric Integrated Brier Score, IBS, could be considered. It is defined as the average Brier Score:
\begin{equation}
    \mathrm{BSS}=\frac{1}{\tau} \int_{0}^{\tau} \mathrm{BS}(t)\mathrm{d} t
\end{equation}
with $\tau$ the study period.

\section{Exposure weighted AUC}
\label{sec:AUC}

When the time of observation is cut into intervals, the problem becomes a binary classification weighted by the exposure. In this case, the weighted AUC is a good measure of the performance of the model. This metric aims at evaluating the ability of the binary classification between dead and alive, when the integration of the \textit{initial exposure} as weight implies giving a bigger importance to the observed individuals rather to the censored ones. The importance given to a mistake on a censored subject increases with the observation time as the information increases as well. It is indeed worse to classify a censored individual as dead if he was observed 90\% of the interval compared to one observed 10\% of it because the first one is less likely to die in the resting time rather than the second one.\\ \\
Let $I_d = \{i : \delta_{i} = 1\}$ and $I_a = \{i : \delta_{i} = 0\}$ be respectively the sets of dead and alive individuals. Considering a threshold function $f$ as follows: 
\begin{equation}
    f_{\tau}(\hat{\delta}) = \left\{
    \begin{array}{ll}
        1 & \mbox{if } \hat{\delta} \geq \tau \\
        0 & \mbox{if } \hat{\delta} < \tau
    \end{array}
\right.
\end{equation}
We then define the two following quantities: \\

\textbf{Weighted true positive rate:}

\begin{equation}
    TPR(\tau) = \frac{1}{\sum_{i\in I_{d}}ei_i}\sum_{i \in I_{d}} \mathbbm{1}_{\{f_{\tau}(\hat{\delta_i}) \neq 0\}} \times ei_i
\end{equation}

\textbf{Weighted false positive rate:}

\begin{equation}
    FPR(\tau) =  \frac{1}{\sum_{i\in I_a}ei_i}\sum_{i \in I_a}  \mathbbm{1}_{\{f_{\tau}(\hat{\delta_i}) = 1\}} \times ei_i
\end{equation}

A weighted ROC curve is drawn by plotting $FPR(\tau)$ and $TPR(\tau)$ for all thresholds $\tau \in \mathbbm{R}$. 

\begin{figure}[H]
    \hspace*{4.3cm}
    \includegraphics[scale=0.4]{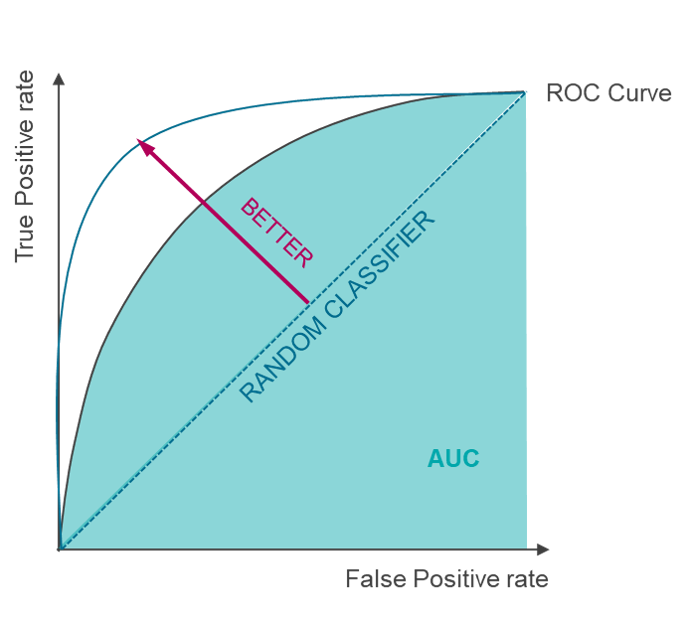}
    \caption{ROC Curve and AUC illustration}
    \label{roc2}
\end{figure}

The weighted AUC is the value of the area under this ROC curve. The AUC ranges from 0.5 to 1 like the C-index, where 1 means that the model is perfect and 0.5 means that the prediction is equivalent to a random classification. \\ \\
The interpretation is quite equivalent to the C-index metric in terms of model quality. However, thanks to the exposures every observation, even the censored ones, may be included to compute the AUC. \\ \\
{\Large\textbf{Survival Modeling}}

Many of the current Machine Learning models have been adapted to survival analysis problems. In this chapter, the theory and the intuition behind the models that have been implemented in the Python library will be given. It is essential to review and understand the theory to make coherent implementation choices. Besides, there is almost no literature about the application of Machine Learning models to discrete data. A deep study of the theory is thus necessary to justify their correct adaptation to predict durations, which is required to derive insurance policy prices. \\ \\
As mentioned before, we considered two approaches to model survival analysis: the models built on the survival dataset and the ones built on the dataset obtained after discretization. In the Python implementation, a main mother class \textit{Model Discrete} and \textit{Model Continuous} has been created for each approach, to gather the common functions of all models (cf Figure~\ref{uml}). These functions are evaluation metrics computation or prediction. Each model, defined in a specific class, finally inherits from the corresponding mother class. As the different implemented models rely on different existing Python packages, such as \textit{statsmodel}, \textit{lifelines} or \textit{scikit-learn}, having a class for each one enables us to deal with all specific constraints. Besides, through the different classes, a homogenization step is included, which contributes to simplifying the library use. Each model can be fitted, evaluated, or can predict mortality thanks to the following pattern: \\ 

\begin{lstlisting}[language=Python]
import scor_survival.models
#Training
model.train(X, event, exposition)
#Evaluation
model.auc(X, event, exposition)
model.ci(X, event, exposition)
#Predicting
model.predict_proba(X)
model.predict_surv(X)
\end{lstlisting}
\section{Time to event framework}

\subsection{Cox-Model adaptation}

Several Machine Learning methods have been adapted to Cox's Proportional-Hazard models such as Trees, Neural-Networks, Generalized Additive Models, etc. In this section, we present Elastic Net and Gradient Boosting Machine adaptation, as they are the most widely used in practice.

\subsubsection{Cox-ElasticNet}

\citet{Tibshirani} proposed to apply the Elastic Net regularization to the Cox proportional hazard model. In the Elastic Net approach, we add a penalization during the parameter estimation process. The goal is to put aside the less relevant features by penalizing the models with a high number of parameters. Decreasing the number of features allows to diminish the signal noise and consequently increase model accuracy.\\
This approach is very useful in high dimensions, i.e. when the number of features is close to the number of observations. This might occur in some epidemiological studies where a significant amount of information is available for each patient, but a limited number of patients is observed.\\
In practice, this approach is often used to quickly identify variables that have the biggest explanatory power and to put aside the non-relevant ones.\\
In this approach, the model parameters $\beta$ are estimated by optimizing the following: 
\begin{equation}
    \hat{\beta}=\underset{\beta}{\operatorname{argmax}}\left[\log( L(\beta) )-\lambda \left ( \alpha \left \|\beta  \right \|_1 +(1-\alpha)\frac{1}{2}\left \|\beta  \right \|_2^2 \right )\right]
\end{equation}
with $L$ the likelihood of the Cox model (cf equation~\ref{coxLL}) and hyper-parameters $\lambda$ and $\alpha$.\\
The penalization intensity is controlled thanks to the hyper-parameter $\lambda$. If $\lambda$ equals 0 then we are performing a classic Cox regression. The higher the value of $\lambda$ the higher the penalization, and the lower the number of non-null parameters.\\
When the hyper-parameter $\alpha$ equals 0, it is called LASSO\footnote{LASSO stands for Least Absolute Shrinkage and Selection Operator} regression, and when $\alpha$ equals 1, it is named Ridge regression. Hyper-parameter $\alpha$ is in $\left[0,1\right]$ and balances between LASSO and Ridge regression.\\
This approach has the same issues as in the classic Cox model. Namely, it heavily relies on the proportional force of mortality assumption that might not be verified. Besides, non-linear effects and unspecified interactions will not be captured by this model.

\subsubsection{Cox - Gradient Boosting Machine}

To integrate non-linear effects within the Cox framework, a gradient boosting adaptation may be considered. \\
Gradient boosting consists in building a complex model thanks to the aggregation of several simple models called weak learners (\citet{Friedman}\&\cite{Friedman2}). The weak learners are all the same base learners throughout the process, but they are successively trained on the residual errors made by the predecessor. Thus, each model relies on previous steps constructed models. \\
Gradient Boosting Machine is a mix between gradient boosting and gradient descent, which is an optimisation process to minimize a loss function. The adaptation of the GBM to a Cox's proportional hazard model~\cite{Ridgeway} is possible by choosing the opposite of Cox partial likelihood as loss function:
\begin{equation}
    LL(\beta) = - \sum _ {i=1} ^ m [X'_{j_{(i)}} \beta - log(\sum _ {j \in R_i} e^{X'_{j_{(i)}} \beta})]
\end{equation}
Generally speaking, the algorithm is presented as the process below: \\ \\
\textbf{\textit{Initialisation}}: $F_{0}(x) = argmin_{\beta} LL(\beta)$ \\ \\
For $m = 1$ to $M$ \textit{(number of weak learners):} 
\begin{itemize}
    \item[$\bullet$] Computation of the pseudo-residuals: $r_{m} = - \frac{dL(F_{m-1}(X))}{F_{m-1}(X)}$
    \item[$\bullet$] Fitting a new weak learner on pseudo-residuals: $f_{m}(X) = r_{m}$
    \item Finding the best $\gamma_m$ by solving $\gamma_m = argmin_{\gamma}L(F_{m-1}(X)+ \gamma \times f_{m}(X))$
    \item[$\bullet$] Update the new model: $F_{m} = F_{m-1} + v\times\gamma_m f_{m}$
\end{itemize}

Thus, at each iteration, until the stopping condition is satisfied, we try to reduce the global error by fitting each specificity of the residuals. A learning rate, $\gamma_m$, is introduced to control how much we adjust the weights of our base learner. This parameter may be constant and chosen at the beginning of the process or optimized at each step. A large value may reduce computing time but may cause divergence, whereas a small one ensures convergence and getting an optimum but makes learning time more consuming. The shrinkage parameter $v$, a scalar between 0 and 1, allows to regularize the model and ensure its convergence.\\
The real advantage of gradient boosting is that it can be adapted to any weak learner. Most of the time, trees are chosen. Cox Gradient Boosting Machine is a way of building classic regression trees by taking into account censoring within the loss function and assuming the proportional hazard hypothesis. In this case, trees are constructed consecutively, and the gradient shows the best path so that each tree is constructed on the previous one in such a way that it leads to the biggest error reduction.   

\begin{figure}[H]
    \centering
    \includegraphics[scale=0.5]{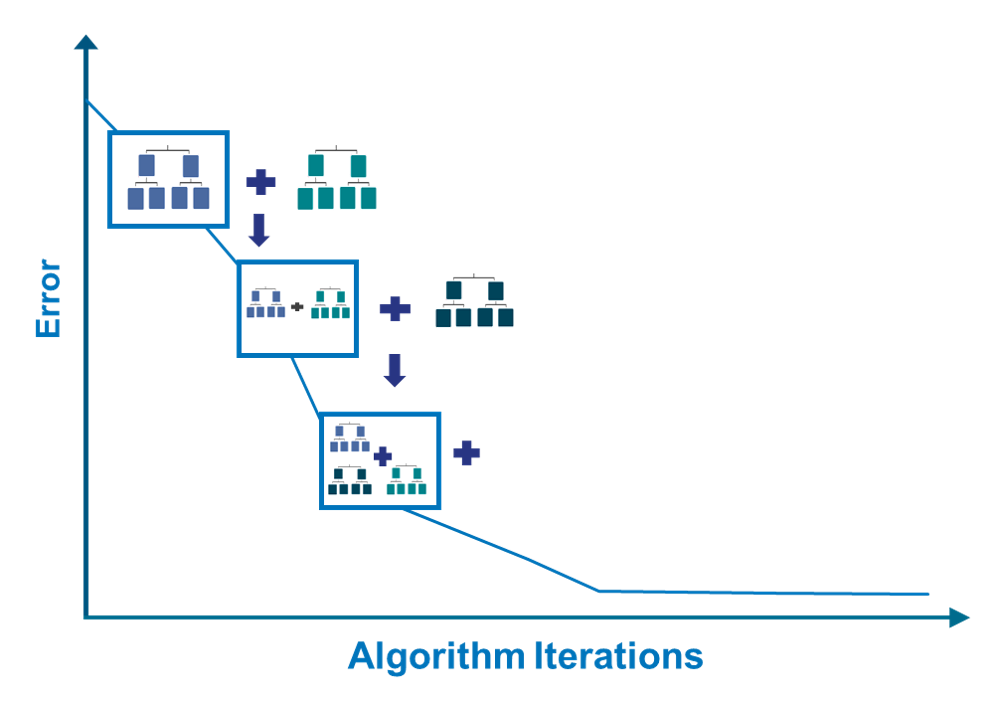}
    \caption{Gradient Tree Boosting}
    \label{BO}
\end{figure}

\subsection{Survival Tree}

Another method to build specific trees for survival analysis has been developed. Compared to Cox-Gradient Boosting, it enables to create predictor trees, which may be directly interpreted. The real advantage of trees is their simplicity compared to other Machine Learning techniques, which contributes to short computation time. \\
Survival trees have been explained by \citet{Bou_Hamad} and \citet{LeBlanc}. It is the direct adaption of decision trees for survival analysis. Traditional decision trees are also called CART (i.e. classification and regression tree), which is the fundamental algorithm of the tree-based methods family. The CART algorithm was developed by Breiman~\cite{CART} and makes the use of trees popular to solve regression and multi-class classification problems. \\ \\ 
Like CART, survival trees are binary trees grown by a recursive splitting of tree nodes. Starting from the root node composed of all the data, the observations are partitioned into two daughter nodes according to a predetermined criterion. Using a recursive process, each node is split again into two nodes until reaching the \textit{stopping criterion}. The best split for a node is found by searching over all possible variables and values, the one that maximizes survival difference.\\
The difference between \textit{CART} and \textit{Survival trees} relies on the \textbf{splitting criterion} used to build the tree. When dealing with survival data, the criterion must explicitly involve survival time and censoring information. Either it aims at maximizing the between-node heterogeneity or at minimizing the within-node homogeneity.

\subsubsection{Log-rank criterion}

The most widely used criterion is the maximization of the log-rank statistic (cf Appendix~\ref{ch:lr}) between the two sub-samples of the nodes, which contributes to creating splits that distinguish the most the mortality. As it is impossible to measure the similarity of the mortality within a group, the idea behind it is that by sequentially creating splits with distinct mortality, we assume we will obtain homogeneous groups at the end as the dissimilar cases become separated at each node. \\
Hyper-parameters should be introduced to optimize the number of splits: a minimum occurrence of events within a leaf or a lower threshold of the log-rank statistic to make a split. The intuition behind these 
\textit{stopping criteria} is to ensure the quality of the split. The first one forces the splitting criterion to be computed on enough data to make sure that the log-rank statistic is consistent. The lower bound for the second one comes from the reject region bound of the underlying log-rank test, which means a node should not be split if the mortality is not statistically different with respect to any variable.\\ \\
The main advantage of this method is that it does not rely on major assumptions to build the tree, even if the statistic considered to measure the difference in mortality between groups is questionable. Indeed, the log-rank test performance may be poor in some situations.\newline
Once the tree is built, the model assumes that individuals within a leaf have the same common survival curve and thus a global survival curve is computed based on the individuals within each final leaves. In open-source packages, the Nelson-Aalen estimator is used to compute the cumulative hazard function, from which we can deduce the survival curve or the expected lifetime duration. Experimental studies have shown that using the Kaplan-Meier estimator to directly estimate the survival curve gives similar results.\\ \\
Thanks to the binary nature of survival trees, individuals with characteristics $x_i$ fall under a unique leaf $f$ composed of observations $(x_{i},\delta_{i})$ with $i \in \mathcal{I}_f$. The prediction of the cumulative hazard function is the estimator for $x_i$'s terminal node:
\begin{equation}
\hat{H}(t|x_i) = \hat{H}_{f}(t) = \sum_{\substack{t_i < t \\ {i \in I_f}}} \frac{d_i}{N_i}\delta_{i}    
\end{equation}

Some other criteria have also been studied such as C-index maximization~\cite{Schmid} or deviance minimization within one node. 

\subsubsection{Deviance criterion}

The deviance minimization is based on a likelihood estimation relying on the proportional hazard function to partition the observations. Under this hypothesis, the hazard function within a leaf $f$ composed of observations $(x_{i},\delta_{i})$ with $i \in \mathcal{I}_f$, is expressed as follows: $$h_{f}(t) = h_{0}(t) \times \theta_f$$
Using the formula~\ref{eq:likelihood}, the likelihood can thus be rewritten as
\begin{equation}
    L = \prod_{f} \prod_{i \in I_f}h_{f}(t_i)^{\delta_i}S_{f}(t_i) = \prod_{f} \prod_{i \in I_f} h_{f}(t_{i})^{\delta_{i}} e^{-H_{f}(t_{i})} = \prod_{f} \prod_{i \in I_f} (h_{0}(t)\theta_f)^{\delta_i}e^{-H_{0}(t_{i})\theta_f}
\end{equation}
Where $H_{0}(t)$ and $h_{0}(t)$ are respectively the baseline cumulative hazard function and the baseline hazard function, and $\theta_{f}$ is the parameter to estimate by likelihood maximisation. When $H_{0}$ is known, we can get the maximum likelihood estimator:
$$\hat{\theta_f} = \frac{\sum_{i \in I_f} \delta_i}{\sum_{i \in I_f} H_0(t_i)}$$
In practice, the cumulative hazard function is unknown and we plug in the Breslow estimator $$\hat{H}_0(t) = \sum_{i: t_i \leq t} \frac{\delta_i}{\sum_{f}\sum_{i: t_i \geq t; i\in I_f}\theta_{f}}$$
The deviance is finally defined as:
\begin{equation}
    R(f) = 2[L_{f}(saturated) - L_{f}(\hat{\theta_{f}})]
\end{equation}
where $L_{f}(saturated)$ is the log-likelihood for the saturated model that allows one parameter for each observation and $L_{f}(\hat{\theta_{f}})$ is the maximal log-likelihood. \\ \\
The algorithm to build the tree adopts the principle of the CART algorithm: it will split the observation and covariate space into regions that maximize the reduction of the deviance realized by the split by testing all possible splits for each of the covariates. In this approach a \textit{stopping criterion} regarding the minimum size of a node is also considered, since the likelihood estimation converges when it relies on a large amount of data. \\ \\
Simulation experiments have shown that the performance is similar to the log-rank statistic. However, this method is not assumption-free and may not be applied to all datasets. \\
According to the trees built with a C-index maximization, the results are quite similar to the ones obtained with trees based on the log-rank statistic and are also assumption-free, but the first ones require much more computation time. Thus, trees using the log-rank criterion should be privileged, which has been shown in several experimental studies.

\subsection{Random Survival Forest}

Random survival forest extends the random forest~\cite{RF} method to right-censored survival data. 

\subsubsection{Random Forest}

Random forest is an ensemble method inspired by the \textit{bagging} of decision trees. Bagging, which means \textit{Bootstrap Aggregating}, is an ensemble learning method that enables the creation of more robust predictors thanks to the aggregation of several ones trained on different subsets.\\
A bagging process consists in generating B random samples with replacement to train B trees $\hat{f}^{b}$ on these subsamples. Finally, the prediction of a new input X is defined as: $$\hat{f}_{Bagging}(X) = \frac{1}{B}\sum_{i=1}^{B} \hat{f}^{b}(X)$$

\begin{figure}[H]
    \hspace*{-1.7cm}
    \includegraphics[scale=0.68]{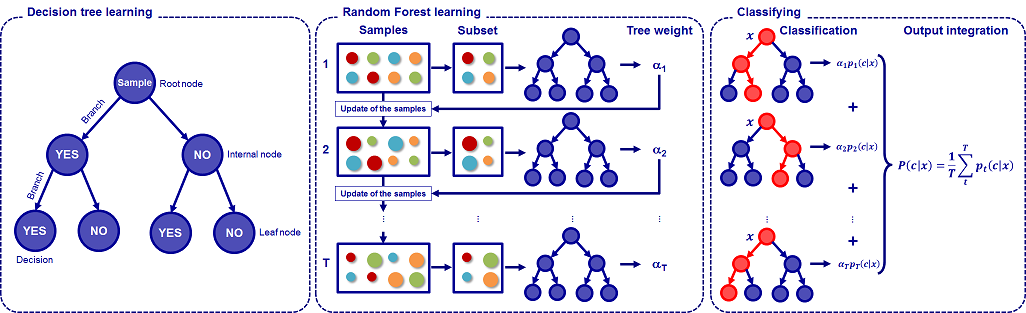}
    \caption{Random forest process illustration (source~\cite{rfill})}
    \label{rf}
\end{figure}

Random forest differs from simple bagging of trees as randomization is not only applied to draw samples but also to select features. At certain nodes, rather than considering all variables, a random subset of the attributes is selected to compute the splitting criterion. The introduction of randomization enables the reduction of the correlation among the trees and the improvement of the predictive performance. The training process is illustrated in Figure~\ref{rf}.

Random survival forest is an ensemble tree method developed by~\citet{Ishwaran} that follows the same process but considers survival trees instead of traditional decision trees. The algorithm is processed as below:

\begin{itemize}
    \item Draw B bootstrap samples from the original data that exclude on average 37\% of the data, called out-of-bag data (OOB data)
    \item Grow a survival tree for each bootstrap sample, by selecting at each node p candidate variables. The split is chosen among the candidate variable that maximizes the survival difference between leaves.
    \item Calculate the cumulative hazard function for each tree, $\hat{H}^{b}(t|x_{i})$ and average it over all the trees to obtain the ensemble cumulative hazard function:
    $$\hat{H}(t|x_{i}) = \frac{1}{B} \sum_{b=1}^{B}\hat{H}^{b}(t|x_{i})$$
\end{itemize}

The interpretation of the result may be questionable as we average several hazard functions to get the predicted one. However, as $H(t) = \int_{0}^{t} h(s)\mathrm{d}s$ is already a sum of functions, averaging it still returns a sum: $\hat{H}(t) = \frac{1}{B} \sum_{b=1}^{B}\int_{0}^{t}\hat{h}^{b}(s|x_{i})\mathrm{d}s = \int_{0}^{t} [ \frac{\sum_{b=1}^{B}\hat{h}^{b}(s|x_{i})}{B}]\mathrm{d}s $ and the prediction makes sense. \\ \\
The main advantage is that forests can model non-linear effects without any prior transformation of the data and, contrary to boosting, in bagging, each tree is built independently and the process can thus be parallelized.

\begin{figure}[H]
    \hspace*{3.2cm}
    \includegraphics[scale=0.6]{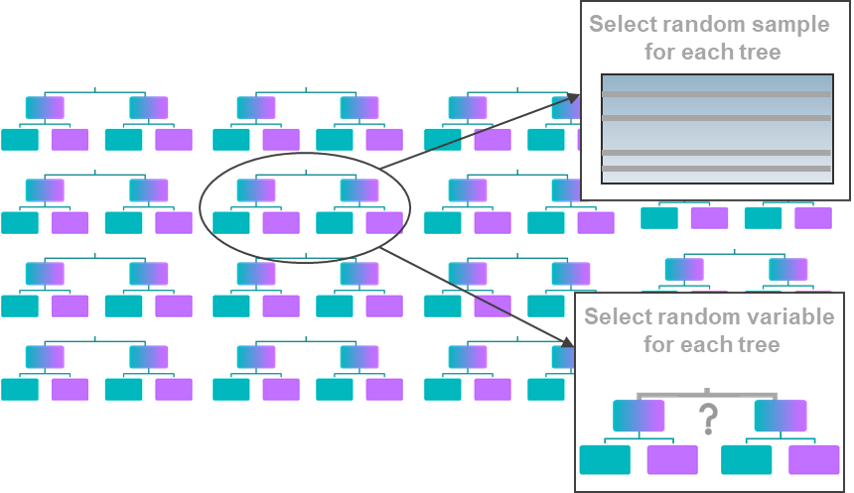}
    \caption{Bagging illustration}
    \label{BA}
\end{figure}

\section{Discrete time modeling}

We will now introduce several models that rely on the pseudo data table and give insights into the use of exposures within modeling. These kinds of models are preferred within the actuarial field, as they may be applied to aggregated population data. Besides, the conversion is only possible from a survival data table into pseudo data tables, which means that only discrete-time modeling is always possible. \\
As sometimes clients directly provide pseudo data tables with only one type of exposure available, it was essential to include this in the Python library models based on the central and initial exposure. \\

\subsection{Poisson regression}

Poisson regression model assumes that the total number of deaths within the time interval $j$ follows a Poisson distribution and is mainly based on the \textbf{central exposure}. That is to say:
\begin{equation}
    d_{j}|X \sim \mathcal{P}(EC_{j}h_{j}(X))
\end{equation}
The idea behind the parameter used in the Poisson distribution comes from the constant hazard function hypothesis, as under this hypothesis $\hat{h}_{j} = \frac{d_{j}}{EC_{j}}$, so that the expected values match.\\
In literature, $d_{j}$ corresponds to an aggregate number of deaths for all similar individuals (i.e. with the same vector $X$ of characteristics for a specific interval). However, due to the additive property of the Poisson distribution, it is equivalent to consider afterwards the aggregation of the prediction of the death indicator of everyone, which means considering a model as follows: 
\begin{equation}
    \delta_{i,j}|X \sim  \mathcal{P}(ec_{i,j}h_{i,j})
\end{equation}
Using a log-link function, the model becomes equivalent to a classic Poisson regression model with the exposure in offset: 
\begin{equation}
    \log(\mathbbm{E[d_{j}|X]}) = \log(EC_{j}) + X'\beta = \log(EC_{j}) + \log(h_{j})
\end{equation}
which means 
\begin{equation}
    \log(h_{j}) = \log(\frac{\mathbbm{E[d_{j}|X]}}{EC_{j}}) = X'\beta
\end{equation}
We then apply the classic generalized linear model with a Poisson distribution to a pseudo data table with exposure. Through the likelihood optimisation with respect to $\beta$, we get the risk parameters: 
\begin{equation}
    L(\beta|X,D,E) = \prod_{j} \frac{(EC_{j}e^{X_j'\beta})^{d_{j}}e^{-EC_{j}e^{X_j'\beta}}}{d_{j}!}
\end{equation}
It is also possible to consider an extension and add a regularisation factor to only consider the variables with a high explanatory power. 
When the probabilities are small, if the \textit{central exposure} is not available in the data, approximating the model with \textit{initial exposure} predicts similar results.\\
Using the maximum likelihood estimator, the hazard function is estimated $\hat{h}_j = \exp(X'\hat{\beta}_j)$, from which the rate of mortality is derived:
$$q_j = 1 - \exp(-\hat{h}_j) = 1 - \exp(-\exp(X'\hat{\beta}_j))$$

\subsection{Binomial regression}

In the case where we try to individually model for each subject the time of death and we aggregate it afterward, one may think of using a logistic regression instead of a Poisson one. Within each interval, $j$, we try to predict the indicator of death $\delta_{i,j}$. Considering a traditional binomial model $\delta_{i,j} \sim B(q_{j})$, where $\delta_{i,j}$ means the death indicator that individual i will die before $\tau_{j}+1$ given he has survived up to $\tau_{j}$ and $q_{j} = P(T \leq \tau_{j} +1| T > \tau_{j})$, will not enable to take censoring into account. \\
Thus we will weigh the model with the \textbf{individual initial exposure}, $ei_{i,j}$, representing the time where i is observed between $[\tau_{j},\tau_{j}+1]$ and equals to 1 if survival exceeds $\tau_{j}$ or if the death is observed. The intuition behind this comes from the Balducci hypothesis to make the estimator matches, as under this assumption  $\hat{q}_{j} = \frac{\sum_{i}\delta_{i,j}}{\sum_{i}ei_{i,j}}$. 
Indeed the weighted likelihood of the model can be expressed as:
\begin{equation}
    L(q_j) = \prod_{i=1}^{l_j} q_{j}^{\delta_{i,j}ei_{i,j}}(1-q_{j})^{(1-\delta_{i,j})ei_{i,j}}
\end{equation}
Then the log likelihood is:
\begin{equation}
l(q_j) = \sum_{i=1}^{l_j} \delta_{i,j}ei_{i,j} log(q_j) + ei_{i,j}(1 - \delta_{i,j})log(1-q_j)
\end{equation}
Deriving the previous equation with respect to $q_j$:
\begin{equation}
    \frac{d l(q_j)}{d q_j} = \sum_{i=1}^{l_j}  \frac{\delta_{i,j}ei_{i,j}}{q_j} - \frac{ei_{i,j} - \delta_{i,j}ei_{i,j}}{1-q_j}
\end{equation}
Thus,
\begin{equation}
    (1-\hat{q}_{j})\sum_{i=1}^{l_j}\delta_{i,j}ei_{i,j} = \hat{q}_{j}\sum_{i=1}^{l_j}ei_{i,j}-ei_{i,j}\delta_{i,j}
\end{equation}
And finally we get the desired estimator $\frac{\sum_{i=1}^{l_j}\delta_{i,j}}{\sum_{i=1}^{l_j}ei_{i,j}} = \hat{q}_{j}$, as $ei_{i,j}$ equals 1 for dead subjects, ie those with $\delta_{i,j}$ equals 1.\\ \\
The previous consideration leads us to consider the \textbf{initial exposure} as a weight in the logistic regression to model the impact of the covariate vector X on the mortality rate. We will thus estimate a risk parameter $\beta$ such as $logit(\mathbbm{E}[{\delta|X}]) = \frac{q(X)}{1-q(X)} = X'\beta$, which means $q(X) = \frac{1}{1+e^{-X'\beta}}$\\ Injecting it in the log-likelihood implies: 
\begin{align*}
    l(\beta) & = \sum_{i,j} \delta_{i,j}ei_{i,j} log(q_{j}(X_{i,j})) + ei_{i,j}(1-\delta_{i,j})log(1-q_{j}(X_{i,j})) \\
    & = \sum_{\substack{{i,j} \\ {\left\{i,j|\delta_{i,j}=0\right\}}}} (-ei_{i,j}log(1+e^{X_{i,j}'\beta}))  + \sum_{\substack{{i,j} \\ {\left\{i,j|\delta_{i,j}=1\right\}}}} (-ei_{i,j}log(1+e^{-X_{i,j}'\beta}))
\end{align*}

If we change the feature space from $\left\{0,1\right\}$ to $\left\{-1,1\right\}$, we can write (by a simple variable change $\delta_{i,j}' = 2\delta_{i,j}-1$) :
\begin{align*}
    l(\beta) & = \sum_{\substack{{i,j} \\ {\left\{i,j|\delta_{i,j}'=-1\right\}}}} (-ei_{i,j}log(1+e^{X_{i,j}'\beta}))  + \sum_{\substack{{i,j} \\ {\left\{i,j|\delta_{i,j}'=1\right\}}}} (-ei_{i,j}log(1+e^{-X_{i,j}'\beta})) \\
    & = \sum_{i,j} (-ei_{i,j}log(1+e^{-\delta_{i,j}'X_{i,j}'\beta}))
\end{align*}
It enables us to get the formula often used in Machine Learning literature and implemented in existing Python packages such as \textit{scikit-learn}, on which our implementation can thus rely. This is then the formula to be maximized. Deriving this equation with respect to $\beta$:
\begin{equation}
    \frac{d l(\beta)}{d \beta} = \sum_{i,j} \frac{ei_{i,j}\delta_{i,j}'X'_{i,j}e^{-\delta_{i,j}'X'_{i,j}\beta}}{1+e^{-\delta_{i,j}'X'_{i,j}\beta}}
\end{equation}
From this equation, we can deduce the first order condition by making it equal to zero. There is no explicit formula, but it can be solved numerically.

\subsection{Generalized Additive Model}

One limit of the two previous models is that only linear interactions are modeled. A first possibility to capture non-linear patterns is to use generalized additive models (GAM). One strength of GAMs is that they produce a regularized and interpretable solution. However, interpretability has a cost as interactions are not detected by the model. Indeed, one should specify the interactions of the variables to be considered. In other words, GAMs strike a nice balance between the interpretable, yet biased, linear model, and the extremely flexible, “black box” learning algorithms.\\ \\
GAMs may be seen as an extension of GLMs. A Generalized Additive Model (semi-parametric GLM) is a GLM where the linear predictor linearly depends on unknown smoothing functions, $s$. Any function could be considered, but in practice, splines are the most widely-used one as they perform well in such circumstances (see \citet{GAM} for more details about GAMs).

\subsubsection{B-splines}

B-splines are curves, made up of sections of polynomials of the degree of the splines, joined together so that they are continuous in value as well as their first and second derivatives. \\
It is a piecewise function built from a polynomial of degree d, where d is a hyper-parameter. To set it, one may use the grid-search method to find the one leading to the best prediction performance. \\
The points at which the sections join are named knots. The locations of the knots must be chosen, for example, the knots would either be evenly spaced throughout the range of observed x values or placed at quantiles of the distribution of unique x values.\\
Finally, B-splines are local functions, that is to say, they are zero everywhere outside the range of their knots.

\begin{figure}[H]
    \centering
    \includegraphics[scale=0.6]{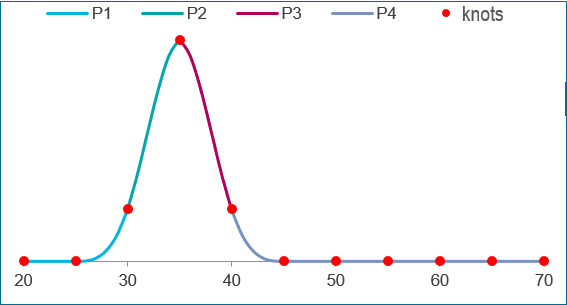}
    \caption{B-splines of degree 3}
    \label{cubsplines}
\end{figure}

Figure~\ref{cubsplines} is an illustration of cubic splines of four polynomials, that is to say $s(x) = P_1(x) + P_2(x) + P_3(x) + P_4(x)$, with polynomials of degree three. The regularity conditions imply at each junction point that the value of the two polynomials and their first and second derivatives are equal. These constraints imply that a spline only has one degree of freedom to be estimates.\\ \\
In the same way, as previously discussed GLMs, Poisson and Binomial, considering exposures allows using GAMs for survival time modeling.

\subsubsection{Poisson GAM}

The previous Poisson regression introduced before will thus be modified to capture the non-linear pattern. The model will still be stated as follows:
\begin{equation}
    d_{j}|X \sim \mathcal{P}(EC_{j}h_{j}(X))
\end{equation}
A log-link function and the use of exposure as offset is kept, however splines are applied to the covariates:
\begin{equation}
    \log(\mathbbm{E[d_{j}|X]}) = log(EC_{j}) + \sum_{k=1}^{p} \theta_{k}S_{k}(X) = \log(EC_{j}) + \log(h_{j})
\end{equation}
which means 
\begin{equation}
    \log(h_{j}) = log(\frac{\mathbbm{E[d_{j}|X]}}{EC_{j}}) = \sum_{k=1}^{p} \theta_{k}S_{k}(X) = S'\theta
\end{equation}
where $\theta = [\theta_1,..,\theta_p]'$ is the vector of regression coefficients and $S$ is the regression matrix, which is the matrix of the B-spline transformation of the covariates: 
$$S_j = [s_1',...,s_n']', \text{ where } s_i' = (S_1(X_{i}),..,S_p(X_{i}))$$
The model then becomes equivalent to the Poisson linear regression if we consider the matrix $S$ instead of the covariates. The $\theta$ parameter is once again obtained through a numerical likelihood maximization.

\subsubsection{Binomial GAM}

Based on the binomial regression section, the \textbf{initial exposure} will be introduced as a weight in the binomial GAM to model the impact of the covariate vector X on the mortality rate. Using the same notations as in the Poisson GAM section, we want to estimate a risk parameter $\theta$ such that $logit(\mathbbm{E}[{\delta|X}]) = \sum_{k=1}^{p} \theta_{k}S_{k}(X) = S'\theta$, which means $q(X) = \frac{1}{1+e^{-S'\theta}}$\\
The problem is once again equivalent to the linear one as long as we consider the matrix of the B-spline transformation of the covariates. The weighted log-likelihood maximization with respect to $\theta$ enables to find an estimator of the risk parameter, from which it is possible to derive the mortality.\\ \\
The estimation of Survival GAM was based on the package \textit{PyGam}. However, this package requires a manual specification of all the variables, their interaction, and the associated spline level. An automatization step had to be introduced compared to the other implemented models to be called in the same manner. 

\subsection{Decision tree}

The binomial regression aims at predicting the death indicator, $\delta$, using initial exposure as weights. As we are predicting a binary variable, it seems interesting to consider a weighted classification problem, such as a weighted decision tree for classification. 

\subsubsection{Classification Tree}

For our purpose, we only need to focus on the intuition behind two-class classification trees. The goal of a binary classification tree is to successively divide observations into two groups with respect to the variable that creates the best split. The partition of the individuals is repeated on each subsample until reaching the \textit{stopping criterion} or having only pure groups at the final step,  that is to say, groups in which all instances have the same label: $\delta$. \\
The algorithm seeks to create at each step two groups as homogeneous as possible by reducing the variance within a group, or equivalently to create two groups as different as possible by increasing the variance between groups. To determine the best split among the features and finding the partition rule, a \textit{splitting criterion} is introduced to measure the quality of a split. In theory, all possible splits should be tested to find the best one, that is to say, the one giving the biggest reduction of variance. However, as it would be time-consuming, some randomization techniques are used in practice.\\
Two criteria are considered to build trees. The first one is the \textit{entropy}, $H(x)$, which aims at measuring the disorder quantity. The second one is the \textit{Gini index}, $G(x)$, which measures the impurity within a group. To make a split from a given node composed of $n_0$ observations $(x_{i},\delta_{i})$ with $i \in \mathcal{I}_0$, we partition our data into two branches $\mathcal{I}_l$ and $\mathcal{I}_r$, with respectively $n_l$ and $n_l$ observations, in order to have the biggest information gain $I(x_0) - [\frac{n_r}{n_0}\times I(x_r)+\frac{n_l}{n_0}\times I(x_l)]$, where $I(x)$ is $G(x)$ or $H(x)$.

\subsubsection{Weighted Tree}

To prevent the bias due to censoring, we need to adapt the CART algorithm by injecting the initial exposure as weights in the \textit{splitting criterion}. \\
Let's define $p_{m}$, the proportion of deaths observed in node $m$, $I_{m}$ the set of individuals of the node with characteristics $x_m$ and $\hat{q}_m$ the death rate estimator for individuals with $x_m$ under the \textit{Balducci hypothesis}: 
$$p_{m} = \frac{1}{\sum_{i \in I_m} ei_{i}} \sum_{i \in I_m} ei_{i}\delta_{i}  = \frac{\sum_{i \in I_m} \delta_{i}}{\sum_{i \in I_m} ei_{i}} = \hat{q}_m \text{ as } ei_{i} = 1 \text{ if } \delta_{i} \neq 0$$
Two splitting criterion may thus be adapted:
\subsubsection*{Entropy:}
$$H(x_m) = - \hat{q}_m\times \log_{2}(\hat{q}_{m}) - (1 - \hat{q}_{m})\times \log_{2}(1-\hat{q}_{m})$$

\subsubsection*{Gini index:}
$$G(x_m) = 2\times \hat{q}_{m}\times (1-\hat{q}_{m})$$

Let's illustrate the intuition behind the Gini index with the ten following individuals within a node m:

\begin{table}[H]
\centering
\begin{tabular}{|c|cccccccccc|}
\hline
 & \textit{\textbf{1}} & \textit{\textbf{2}} & \textit{\textbf{3}} & \textit{\textbf{4}} & \textit{\textbf{5}} & \textit{\textbf{6}} & \textit{\textbf{7}} & \textit{\textbf{8}} & \textit{\textbf{9}} & \textit{\textbf{10}} \\ \hline
\textit{\textbf{$\delta$}} & 1 & 1 & 1 & 1 & 1 & 0 & 0 & 0 & 0 & 0 \\
\textit{\textbf{ei}} & 1 & 1 & 1 & 1 & 1 & 1 & 1 & 0.2 & 0.8 & 0.5 \\ \hline
\end{tabular}
\label{ind}
\end{table}

Using the formulas, we have: $\hat{q}_m = \frac{5}{8.5} = 0.59$ and $G = 2\hat{q}_m(1-\hat{q}_m) = 0.48$. 

\subsubsection{First split}

Considering a perfect split where we have on the left side the first 5 individuals and on the right side the 5 others:
$\hat{q}_l = 1$ and $\hat{q}_r = 0$ thus $G_l = 0$ and $G_r = 0$ \\
The information gain is thus $I = 0.48$

\subsubsection{Second split}

Considering a variable that splits on the left side the dead and censored individuals and on the right the survivors: $\hat{q}_l = 0.77$ and $\hat{q}_r = 0$ thus $G_l = 0.36$ and $G_r = 0$ \\
The information gain is thus $I = 0.20$

\subsubsection{Third split}

Considering a variable that splits on the left side the dead and alive individuals and on the right the censored ones: $\hat{q}_l = 0.71$ and $\hat{q}_r = 0$ thus $G_l = 0.40$ and $G_r = 0$ \\
The information gain is thus $I = 0.19$

\begin{figure}[H]
    \hspace*{-1.7cm}
    \includegraphics[scale=0.6]{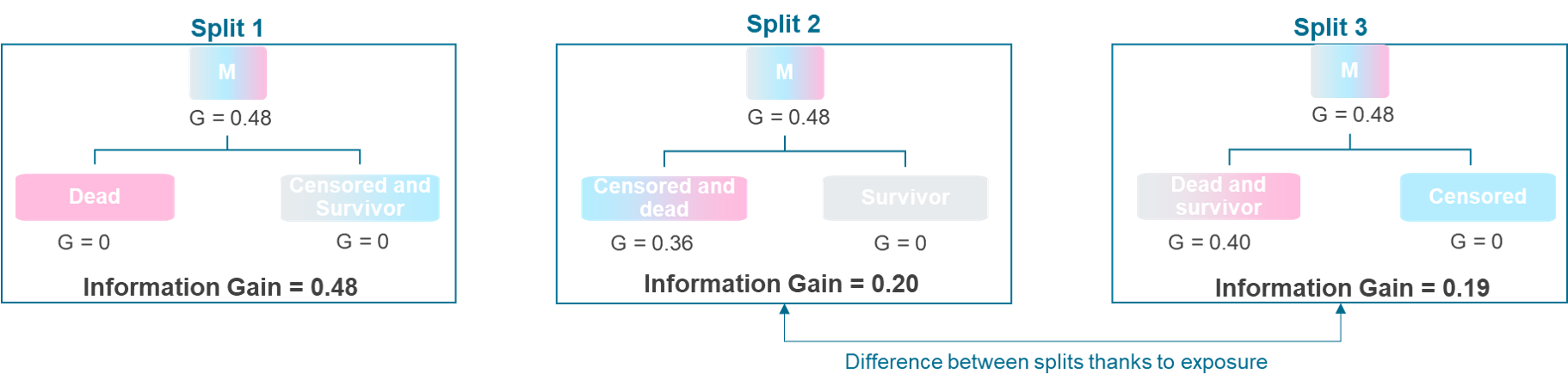}
    \caption{Split representation}
    \label{te}
\end{figure}

Thus the information gain is maximal for the first split as it creates two pure leaves.\\ As shown with the second and the third splits, censoring is indeed taken into account thanks to the exposure. Without adjusting the probability with the initial exposure, both splits would have been equivalent (as it becomes equivalent to consider all exposures equal to one and thus there is no difference between censored and alive people for the Gini index). \\
However, in our case, mixing censored and dead observations within a leaf is better than mixing alive and dead ones. This can be explained as the Gini index increases if we reduce the global exposure, keeping the number of dead observations the same and as the dominant label of the leaf. We can deduce the same conclusion on the ascendant side of the Gini curve by interchanging dead and alive observations. \\ 
Censored and dead VS alive splits are indeed preferable compared to alive and dead VS censored. For censored people we know for sure that they survive until a certain point of time. However the probability of dying before the end of the observation period is not null compared to the survivors. Thus the mistake is less certain. 

\begin{figure}[H]
    \centering
    \includegraphics[scale=0.4]{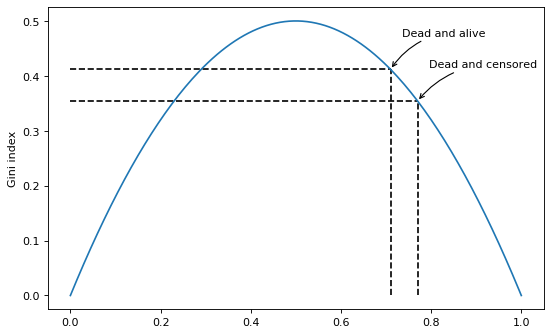}
    \caption{Impurity curve}
    \label{gindex}
\end{figure}

One may also think of building a tree based on the deviance criterion such as described for the survival tree. In this case, the weighted log-likelihood of the binomial regression should be considered to measure the deviance:
$$R(x_m) = 2[L_{m}(saturated) - L_{m}(\hat{\beta_{m}})] = 2\sum_{i\in I_m} ei_{i}log(\frac{1 + e^{-\delta_{i}X'_{i}\hat{\beta}_{m}}}{1 + e^{-\delta_{i}X'_{i}\beta_{s}}})$$
From this definition of the tree, a random forest algorithm may be derived as explained in the time-to-event section.

\subsection{Gradient boosting}

Using the Balducci hypothesis, we proved that the probability of dying in a specific interval j can be estimated as the ratio of the death indicator within this interval on the \textbf{initial exposure}:
\begin{equation}
 P(T \leq \tau_{j}+1 | T_{i} > \tau_{j}) = \frac{\sum_{i}^{l_{j}}\delta_{i,j}}{\sum_{i}^{l_{j}}ei_{i,j}}
\end{equation}

From this result and the previous considerations on the logistic regression, we will thus consider traditional Machine Learning models to predict the probability of death within a time interval as the problem may become equivalent to a binary classification of the death indicator as long as \textbf{initial exposures} are used to weigh the loss function in the model. We will focus on weighted-Gradient Boosting Machine (XGBoost, CatBoost, LightBoost).\\ \\
\textbf{\textit{Initialisation}}: $F_{0}(x) = argmin_{\beta} LL(\beta)$ \\ \\
For $m = 1$ to $M$ \textit{(number of weak learners):} 
\begin{itemize}
    \item[$\bullet$] Computation of the pseudo-residuals: $r_{m} = - \frac{dL(F_{m-1}(X))}{F_{m-1}(X)}$
    \item[$\bullet$] Fitting a new weak learner on pseudo-residuals: $f_{m}(X) = r_{m}$
    \item Finding the best $\gamma_m$ by solving $\gamma_m = argmin_{\gamma}L(F_{m-1}(X)+ \gamma \times f_{m}(X))$
    \item[$\bullet$] Update the new model: $F_{m} = F_{m-1} + v\times\gamma_m f_{m}$
\end{itemize}

Classification trees as used as weak learners, $f_m$ and the opposite of the weighted-log-likelihood of the binomial regression as the loss function: 
\begin{equation}
   L(\beta) = \sum_{i} \delta_{i}ei_{i} log(q(X_{i})) + ei_{i}(1-\delta_{i})log(1-q(X_{i}))
\end{equation}
Where  $q(X) = \frac{1}{1+e^{-X'\beta}}$

\subsubsection{XGBoost}

XGBoost (short for extreme gradient boosting) is an implementation of gradient boosted decision trees designed for speed and performance. It is initially created by~\citet{XGBoost} and is now maintained by many developers. It offers a flexible framework for tree-based gradient boosting with randomly sub-sampling and regularization. But the most attracting part of this method is that its training speed and model robustness are much better than traditional gradient boosting.

Using the original notation of the XGBoost paper~\cite{XGBoost}, at time t, we have:
\begin{equation}
    L^{(t)} = \sum_{i=1}^{n} l(y_i, \hat{y_i}^{(t-1)} + f_t(x_i)) + \Omega(f_t) \approx \sum_{i=1}^{n} l(y_i, \hat{y_i}^{(t-1)}) + g_i f_t(x_i) + \frac{1}{2} h_i f_t(x_i)^2 + \Omega(f_t)
\end{equation}

In our case, we are interested at examining the gradient boosting for:
\begin{equation*}
    l(y_i, \hat{y_i}^{(t-1)} + f_t(x_i)) = ei_i y_i log(\frac{1}{1+e^{-\hat{y_i}^{(t-1)- f_t(x_i)}}}) + ei_i (1-y_i) log(1-\frac{1}{1+e^{-\hat{y_i}^{(t-1)} - f_t(x_i)}})
\end{equation*}

The gradient
\begin{align*}
    g_i & = \frac{\partial l(y_i, \hat{y_i}^{(t-1)})}{\partial \hat{y_i}^{(t-1)}}\\
    & = -ei_i y_i \frac{e^{-\hat{y_i}^{(t-1)}}}{1 + e^{-\hat{y_i}^{(t-1)}}} + ei_i(1-y_i) \frac{e^{\hat{y_i}^{(t-1)}}}{1 + e^{\hat{y_i}^{(t-1)}}}\\
    & = -ei_i y_i \frac{e^{-\hat{y_i}^{(t-1)}}}{1 + e^{-\hat{y_i}^{(t-1)}}} + ei_i(1-y_i) \frac{1}{1 + e^{-\hat{y_i}^{(t-1)}}}\\
    & = -ei_i y_i + ei_i \frac{1}{1+e^{-\hat{y_i}^{(t-1)}}} \\
    & = ei_i \hat{q_x}^{(t-1)}(x_i) - ei_i y_i
\end{align*}
With the log likelihood as the loss function, we see that the gradient for each instance $i$ is proportional to the difference between the predicted dead and the observed ones weighted by the initial exposure. More importance is given to the observations with a bad mortality prediction.

The hessian
\begin{equation}
    h_i = \frac{\partial^{2} l(y_i, \hat{y_i}^{(t-1)})}{\partial \hat{y_i}^{(t-1)}} = ei_i \frac{e^{-\hat{y_i}^{(t-1)}}}{(1+e^{-\hat{y_i}^{(t-1)})^2}} =  ei_i \hat{q_x}^{(t-1)}(x_i) (1 - \hat{q_x}^{(t-1)}(x_i))
\end{equation}

The optimal weight
\begin{equation}
    w_j^* = - \frac{\sum_{i \in I_j} g_i}{\sum_{i \in I_j} h_i + \lambda} = - \frac{\sum_{i \in I_j} ei_i \hat{q_x}^{(t-1)}(x_i) - ei_i y_i}{\sum_{i \in I_j} ei_i \hat{q_x}^{(t-1)}(x_i) (1 - \hat{q_x}^{(t-1)}(x_i)) + \lambda}
\end{equation}

The value of the weight $w_j^*$ can be high and the initial second-order Taylor approximation can thus be no longer sustainable. A shrinkage factor $\eta$ can be introduced here to diminish the importance of the current learner.\\

The scoring function for the weak learner t is
\begin{equation}
    L^t(q) = - \frac{1}{2} \sum_{j=1}^T  \frac{(\sum_{i \in I_j} ei_i \hat{q_x}^{(t-1)}(x_i) - ei_i y_i)^{2}}{\sum_{i \in I_j} ei_i \hat{q_x}^{(t-1)}(x_i) (1 - \hat{q_x}^{(t-1)}(x_i)) + \lambda} + \gamma T
\end{equation}

The literature also recommends using some subsampling when constructing the tree to diminish the greediness of the trees.\\
In a word, the XGBoost corrected with the initial exposure is a model that will correctly integrate the partial information for each duration where an individual is observed. The learners recursively segment the space with weights dependant on the performance of the precedent learners and the regularisation parameters. This dependency leads to difficulties in the analysis of the structure of the trees. However, the celebrity of the XGBoost is established and the model has demonstrated high performances in many data science competitions.

\subsubsection{CatBoost}

CatBoost is an open-source Machine Learning algorithm developed by Yandex~\cite{CatBoost} that uses gradient boosting on decision trees. The difference with other gradient boosting algorithms is that it successfully handles categorical features and uses a new schema for calculating leaf values when selecting the tree structure, which helps to reduce overfitting. Most popular implementations of gradient boosting use decision trees as base predictors. However, decision trees are convenient for numerical features, but, in practice, many datasets include categorical features, which are also important for prediction. Categorical features are not necessarily comparable with each other and most of the time, to deal with categorical features in gradient boosting, a pre-processing step is needed to convert the categorical values into numbers before training. With Catboost, this step is no longer needed but the categorical features cannot contain missing values. \\
Within the developed library, a pre-processing step is thus still implemented to replace all missing values with the label \textit{'Missing value'}, when some are contained in a categorical variable. 

\subsubsection{LightGBM}

Microsoft researchers~\cite{LightGBM} have proposed a novel open-source gradient boosting algorithm called LightGBM. This new implementation aims at optimizing the process for large datasets with a high feature dimension. Contrary to other gradient boosting implementations, LightGBM does not scan all the data instances to estimate the information gain of all possible split points thanks to two novel techniques: \textit{Gradient-based One-Side Sampling} and \textit{Exclusive Feature Bundling} to deal with a large number of data instances and a large number of features respectively.\\
The principle of \textit{Gradient-based One-Side Sampling} is to exclude a significant proportion of data instances with small gradients as they play a less important role in the information gain, and only use the rest to estimate the information gain. The principle of \textit{Exclusive Feature Bundling} is to reduce the number of features by mutually bundling exclusive features (i.e., they rarely take nonzero values simultaneously). 
The experimental results show that LightGBM can significantly outperform XGBoost in terms of computational speed and memory consumption while achieving almost the same accuracy. \\ \\

{\Large \textbf{Interpretation}}

Some of the above-presented models are \textit{"black box"} Machine Learning models, which cannot be directly interpreted. However, in many cases, understanding the model remains essential for different reasons: respecting the regulatory requirement to justify every decision taken, maintaining stakeholder trust by understanding the produced results, etc.\\
For that purpose, three complementary methods of interpretation have been implemented within the library. The presented methods are post-hoc interpretation methods, i.e. once the model has been fitted, and agnostic to the model, i.e. independent of the algorithm to be explained (see~\cite{ddint} for more details).

\section{Permutation variable importance}
\label{sc:pi}

Feature importance can be assessed using the permutation method which has been introduced by~\citet{PFI}. In this method, we measure the increase in the prediction error of the model after we permuted the values of the features. The permutation breaks the relationship between the feature and the true outcome thus it increases the prediction error. The higher the increase in the prediction error, the higher the importance of the feature.\\
In practice, after having trained a model and assessed the model performance, for each feature $j$ one can proceed as follows:
\begin{itemize}
    \item For each observation, replace the feature with a randomly generated variable. This enables to break the relationship between the feature and the true outcome to be predicted.
    \item Produce model forecast and assess the loss in model performance $L^{j}$. It can either be the difference or the ratio between the original error of the model and the one after permutation. 
\end{itemize}
A high loss in performance shows that the model heavily relied on the feature to produce the predictions. If performances are unchanged after shuffling values of one feature, it means that the model ignored the feature for the prediction. Thus, the importance of the features is ranked by their descending $L^{j}$.\\ \\
This approach provides a global insight into model behavior. By permuting a feature, we do not only break the relationship with the outcome variable but also the interaction effects with all the other variables. The inclusion of the interaction effect within the measure not only has advantages but also disadvantages. The importance of the interaction is indeed measured twice, in the two associated features. \\
The big advantage of this approach is that it is less time-consuming, as we do not need to retrain the model. This method only requires the application of the trained model to different observations and an error computation. \\ \\
Within the library, every model class inherits a permutation variable importance method. This method shuffles the modalities of a variable and assesses the loss in the performance of the model on the noisy data.

\section{Partial Dependence}
\label{sc:pd}

This method is the oldest one to interpret Machine Learning models and has been introduced by~\citet{Friedman}. The partial dependence plot shows the marginal effect that one or two features have on the predicted outcome. In practice, more features could be considered but we limit to one or two for representation purposes, i.e. producing 2D or 3D plots. The plot can show whether the relationship between the target and a feature is linear, monotonic, or more complex.\\
The \textit{Partial Dependence function} is defined as the average model prediction for a given value of a feature. When dealing with numerical features, it can be defined as follows:
$$\hat{f}_{x_s}(x_s) = \mathbbm{E}_{x_c}[\hat{f}(x_s,x_c)] = \int \hat{f}(x_s,x_c)\partial\mathbbm{P}(x_c) $$
where $x_s$ are the features for which the Partial Dependence function should be plotted and $x_c$ the other features used to train the model. In practice, there are one or two features in the set $S$, those for which we want to know the effect. The \textit{Partial Dependence function} is estimated thanks to:
$$\hat{f}_{x_s}(x_s) = \frac{1}{n} \sum_{i=1}^{n}\hat{f}(x_s,x_c^{(i)})$$
where $x_c^{(i)}$ are feature values from the data for the features for which we are not interested in measuring the impact.\\ \\
To produce the Partial Dependence curve in practice, after having trained a model, one can process as follows:
\begin{itemize}
    \item Consider a data sample
    \item Select a feature
    \item Replace the feature values by x
    \item Compute the average model prediction for each value x of the feature thanks to the previous formula
\end{itemize}

When dealing with categorical features, the computation of the Partial Dependence is a bit different. For each of the categories, we get an estimate by forcing all data instances to have the same category. For example, if we are interested in the Partial Dependence plot for gender, we get 2 numbers, one for each. To compute the value for "female", we replace the gender of all data instances with the number associated with "female" and average the predictions.\\ \\
If the computation of this kind of plot is intuitive and has a causal interpretation, one has to keep in mind that Partial Dependence only gives the average trend. It can thus differ a lot when looking at subsets. Besides, the Partial Dependence plot relies on the assumption that the features in C and S are not correlated. When the assumption holds, the \textit{Partial Dependence Plot} perfectly represents how the feature influences the prediction on average. That is to say how the average prediction in the dataset changes when the j-th feature is changed. However, if this assumption is violated the averages computed for the Partial Dependence plot will include data points that are very unlikely such as a two-meters tall person weighing thirty kilograms. \\ \\
Every model class implements a Partial Dependence function. For discrete models, mortality is predicted for different modalities of the variable. For survival models, the hazard rate is the indicator. The models unable to handle categorical features transform the data after the variable has been set to a particular modality.

\section{SHAP}

SHAP (\textit{SHAPley Additive exPlanations}) by~\citet{SHAP} is a method to explain individual predictions. The goal of SHAP is to explain the prediction of an instance x by computing the contribution of each feature to the prediction. It enables to quantify the role of each feature in the final decision of a model. \\
The principle comes from the game theory as we consider the prediction as the payout of a game in which each feature value of the instance is a "player". SHAP will then be based on the game theoretically optimal Shapley values as these values tell us how to fairly distribute the "payout" (i.e. the prediction) among the features. \\ \\
Let's explain the general idea thanks to an example inspired by the one given in his book by~\citet{Interpretation}. We have trained a Machine Learning model to predict a life duration. For a certain individual, it predicts 10 years and we need to explain this prediction. The individual is a 65-year-old male and a current smoker with a past cancer history. The average prediction for all individuals is a remaining life of 8 years. SHAP tries to explain how much each of the previous feature values have contributed to the prediction compared to the average prediction thanks to the game theory intuition. \\
The "game" is to predict every single instance. The "players" are the feature values of the instances, in our example, they are the gender, age, smoking habits, and cancer history. Together, these features implied a prediction of a 10-year life duration when the average prediction is 8 years. The goal is then to explain the difference of -2 years between the two. For example, the answer could be: the age contributed 15 years, the gender contributed 1 year, the smoking situation contributed -10 years, and the past cancer history -8 years. The contributions add up to -2 years of remaining life, which is the final prediction minus the average predicted remaining life duration. \\ \\
Concretely, the \textit{Shapley value} is the average marginal contribution of a feature value across all possible coalitions. We will evaluate the contribution of the age when it is added to a coalition of gender and smoking status. We simulate that only these three features are in a coalition by randomly drawing another individual from the data and using its value for the cancer history. The value of cancer history is replaced by the randomly drawn, let's assume it is no past cancer. Then we predict the remaining life duration of an individual with this combination, assuming the model predicts $x_1$ years.\\ 
In a second step, we remove the age from the coalition by replacing it with the age value from the randomly drawn individual which may be the same or different as it is randomly drawn. We predict again the duration with this new age value and the prediction becomes $x_2$. Thus, the contribution of the new age was $x_1 - x_2$. This estimate strongly depends on the values of the randomly drawn observation that served as a "donor" for the cancer history and a feature value. To get the best estimate as possible, the idea is to repeat this sampling step, averaging all the contributions.\\ \\
Formally to approximate the Shapley estimation for a single feature value j of the instance of interest x, one can thus iterate M times the following m process:
\begin{itemize}
    \item Draw random instance $z$ from the data matrix $X$
    \item Choose a random permutation o of the feature values
    \item Order instance x: $x_{o} = (x_{(1)},...,x_{(j)},...,x_{(p)})$
    \item Order instance z: $z_{o} = (z_{(1)},...,z_{(j)},...,z_{(p)})$
    \item Construct two new instances:
    \begin{itemize}
      \item With feature j: $x_{+j} = (x_{(1)},...,x_{(j-1)},x_{(j)},z{(j+1)}...,z_{(p)})$
      \item Without feature j: $x_{-j} = (x_{(1)},...,x_{(j-1)},z_{(j)},z{(j+1)}...,z_{(p)})$
    \end{itemize}
    \item Compute the marginal contribution $\phi_{j}^{m}(x) = \hat{f}(x_{+j}) - \hat{f}(x_{-j})$
\end{itemize}
The Shapley value is the average over the M iterations: $\phi_{j}(x) = \frac{1}{M}\sum_{m=1}^{M} \phi_{j}^{m}(x)$\\
The interpretation of the Shapley value for feature value j, $\phi_{j}$, is: given the current set of feature values, the contribution of the value of the j-th feature is $\phi_{j}$ to the difference between the actual prediction of this particular instance and the average prediction for the data.\\
Then the procedure must be repeated for each of the features j, to get all Shapley values.\\
SHAP is thus quite time-consuming as this process has to be repeated for all possible coalitions. Computation time increases exponentially with the number of features, that is why most of the time only the contributions for a few numbers of samples of the possible coalitions are computed.

\clearpage

\newpage
\appendix
{\Large \textbf{NHANES variables dictionary}}
\label{ch:NHANES}
\begin{itemize}

\subsection*{Target variables}

\item \textbf{Surv}: followed-up time in month
\item \textbf{Death}: status of mortality at the end of observation

\subsection*{Categorical variables}

\item \textbf{Gender}: male or female
\item \textbf{Alcohol}: regular alcohol consumer?
\item \textbf{Smoker}: regular  smoker?
\item \textbf{Marital status}: marital status, in 6 categories : \textit{Never Married/ Married/ Living With Partner/ Separated/ Divorced/ Widowed}
\item \textbf{Educational Level 20plus}: education level for individual more than 20 years old, in 5 categories : \textit{High School Grad GED or equivalent/ Some College or AA Degree/ College Graduate or Above/ 9-11th Grade/ Less Than 9th Grade}
\item \textbf{Health insurance coverage}: covered by a health insurance?
\item \textbf{general health condition}: self-evaluation of one’s health status, in 6 categories : \textit{Good/ Very Good/ Fair/ Excellent/ Poor}
\item \textbf{Tot Income family}: total family income, on income amount category
\item \textbf{pastCancer}: has cancer history? 
\item \textbf{compare activity same age}: self-evaluation for activity level compared with peers, in 3 categories \textit{As Active/ More Active/ Less Active}
\item \textbf{Sleep hours}: daily average sleep hours, on hours category

\subsection*{Numerical variables}

\item \textbf{Age}: one’s age when he is interviewed
\item \textbf{BMI}: body mass index
\item \textbf{Weight body metric}: body weight
\item \textbf{STEPS}: The step count recorded by the physical activity monitor
\item \textbf{Family Poverty income ratio}: ratio of family income to poverty threshold
\item \textbf{Glycohemoglobin}: glycohemoglobin \%
\item \textbf{plasGluMG}: fasting glucose (mg/dL)
\item \textbf{SBP}: systolic blood pressure
\item \textbf{TotalCholesterol}: total cholesterol (mmol/L)
\item \textbf{LDL}: LDL cholesterol (mmol/L)
\item \textbf{Cotinine}: cotinine by serum test (ng/mL)
\item \textbf{Pulse}: 60s heart pulse
\item \textbf{HDL}: HDL cholesterol (mmol/L)
\item \textbf{total fat monosat}: total monounsaturated fatty acids (gm)
\item \textbf{Waist circumference}: waist circumference
\item \textbf{total fat}: total fat (gm)
\item \textbf{DBP}: diastolic blood pressure
\end{itemize}

{\Large \textbf{Kaplan-Meier Estimator}}
\label{ch:KM}

The main idea behind this estimator is that surviving after a given time $t$ means being alive just before $t$ and do not die at the given time $t$. Consequently, with $t_{0} < t_{1} < t $ we get : \\
\begin{align*}
S(t) & = \mathbb{P}(T > t) \\
 & = \mathbb{P}(T > t_{1} \ , \ T > t) \\ 
 & = \mathbb{P}(T > t \ | \ T > t_{1}) \times \mathbb{P}(T > t_{1}) \\ 
 & = \mathbb{P}(T > t \ | \ T > t_{1}) \times \mathbb{P}(T > t_{1} \ | \ T > t_{0}) \times \mathbb{P}(T > t_{0})\\ 
\end{align*}

In the end, by considering all the distinct times $t_{i}, (i=1,...,n) $ where an event occurred ranked by increasing order (whatever it is a death or censorship) we get: \\
$ S(t_{j}) = \mathbb{P}(T > t_{j}) = \prod\limits_{i=1}^{j}\mathbb{P}(T > t_{i} \  | \  T > t_{i+1}) $ , with $t_{0}=0 $.\newline 

Considering the following:
\begin{itemize}
	\item[$d_{j}$ ] the number of deaths that occurred in $t_{j}$
    \item[$N_{j}$] the number of individuals alive just before $t_{j}$
\end{itemize}

The probability $q_{j} = \mathbb{P}(T \leq t_{j} \ | \ T > t_{j-1}) $ of dying in the time interval $]t_{j-1},t_{j}]$ knowing the individual was alive in $t_{j-1}$ can be assessed by : $\hat{q_{j}}=\frac{d_{j}}{N_{j}}$ \\
Let $\delta_{i}$ be the censorship indicator of each observation; the Kaplan-Meier estimator is then defined as:
\begin{equation}
\hat{S}(t)=\prod \limits _{t_{i}\leq t}(1- \frac{d_{i}}{N_{i}})^{\delta_{i}} 
\end{equation}

\begin{Large} \textbf{Cox likelihood} \end{Large}
\label{ch:Cox}

Cox model is defined as $ h(t \ | \  X) = h_{0}(t) \times exp(\beta' X)$. \\
Let consider 3 individuals A,B and C to give the intuition behind the formula. Considering that there is no tie and that $t_{(i)}$ is the ranked sequence of death time, for $i \in [1;3]$.\\
Let $R_i$ be the risk set, which is a set of indices of the subjects that are still alive just before $t_{(i)}$ : $R_i = \{ j: t_j \leq t_{(i)} \}  = \{A,B,C \}$ \\
The contribution to the likelihood will be the conditional probability :
$$\mathbbm{P}(\text{what happened at~} t_{(i)} \text{| one event occurs at } t_{(i)} \text{and the information up to~} t_{(i)})$$
Considering without loss of generality that A dies at $t_{(i)}$ :
\begin{align*}
 \mathbbm{P_{t_i}} & = \mathbbm{P}(\text{A died; B,C survived | A,B,C in the risk set and one died}) \\ \\
 & = \frac{\mathbbm{P}(\text{A died; B,C survived})}{\mathbbm{P}(\text{A died; B,C survived}) + \mathbbm{P}(\text{B died; A,C survived}) + \mathbbm{P}(\text{C died; A,B survived})}
\end{align*}
The events being independent using the previous notation:
\begin{align*}
 \mathbbm{P_A} & = \mathbbm{P}(\text{A died; B,C survived}) \\ 
 & = \mathbbm{P}(t_A = t_{(i)}, t_B > t_{(i)}, t_C > t_{(i)}) \\
 & = f_{A}(t_{(i)})\times S_{B}(t_{(i)}) \times S_{C}(t_{(i)}) \\
 & =  h_{A}(t_{(i)})\times S_{A}(t_{(i)}) \times S_{B}(t_{(i)}) \times S_{C}(t_{(i)}) \\
 & =  h_{A}(t_{(i)})\times C \text{      where  } C = S_{A}(t_{(i)}) \times S_{B}(t_{(i)}) \times S_{C}(t_{(i)})
\end{align*}

Injecting $P_A$ in $P_{t_i}$ and deducing the same formula for B and C :
\begin{align*}
 \mathbbm{P_{t_i}} & = \frac{h_{A}(t_{(i)})\times C}{h_{A}(t_{(i)})\times C + h_{B}(t_{(i)})\times C + h_{C}(t_{(i)})\times C} \\ \\ 
 & = \frac{h_{0}(t_{(i)}) \times exp(\beta' X_{A})}{h_{0}(t_{(i)}) \times exp(\beta' X_{A}) + h_{0}(t_{(i)}) \times exp(\beta' X_{B})+ h_{0}(t_{(i)}) \times exp(\beta' X_{C})}
\end{align*}

Finally we find :
$$P_{t_i} = \frac{exp(\beta'X_{A})}{\sum_{j \in R_{i}} exp(\beta'X_{j})}$$
The partial likelihood is then the multiplication over all the time of events:
$$L = \prod_{i=1}^{3} P_{t_i} = \prod_{i=1}^{3} \frac{exp(\beta'X_{j_{(i)}})}{\sum_{j \in R_{i}} exp(\beta'X_{j})}$$
The general formula for n individuals is obtained with the same method. 

{\Large \textbf{Logrank statistics}}
\label{ch:lr}

The logrank test is the most widely used test to compare survival curve. It is a non-parametric test. This test can be generalized for any number of groups but for the survival tree purpose only the two groups version is used. 
\\ \\
In this test, we state the null hypothesis :
$H_{0} : S_{A}(t) = S_{B}(t) \text{          } \forall t$

Within each group, we compute for each observed time i, the number of expected death:
$$e_{Ai} = \frac{n_{Ai}d_{i}}{n_{i}} \text{ and }  e_{Bi} = \frac{n_{Bi}d_{i}}{n_{i}}$$

We finally aggregate for each time to obtain the total number of expected death $E_{.} = \sum_{i} e_{.i}$ and the observed number of death $O_{.} = \sum_{i} d_{.i}$

\begin{table}[H]
\centering
\begin{tabular}{|c|ccc|}
\hline
 & \textbf{Group A} & \textbf{Group B} & \textbf{Total} \\ \hline
\textit{Death} & $d_{Ai}$ & $d_{Bi}$ & $d_{i}$ \\
\textit{Survivorship} & $n_{Ai} - d_{Ai}$ & $n_{Bi} - d_{Bi}$ & $n_{i} - d_{i}$ \\
\textit{Total} & $n_{Ai}$ & $n_{Bi}$ & $n_{i}$ \\ \hline
\end{tabular}
\caption{Notation used for the number at time i}
\label{lr}
\end{table}

Finally the logrank statistics is given by $X^{2} = \frac{(O_{A} - E_{A})^{2}}{E_{A}} + \frac{(O_{B} - E_{B})^{2}}{E_{B}}$ \\
Under $H_{0}$ this statistics follows a chi square distribution with one degree of freedom, we can then compute the p-value. \\ \\
As the whole survival theory, the validity of this test relies on the independence assumption between the observed event and the censoring. \\
The main limit is the difficulty to reveal the difference of mortality between two groups when their survival curve cross. The power of the test is indeed maximum for proportional curves.

\end{document}